# BF-GAN: Development of an AI-driven Bubbly Flow Image Generation Model Using Generative Adversarial Networks

Wen Zhou[1], Shuichiro Miwa[1*], Yang Liu[2], Koji Okamoto[1]

[1]Department of Nuclear Engineering and Management, School of Engineering, The University of Tokyo, 7-3-1 Hongo, Bunkyo-ku, Tokyo 113-8654, Japan
[2] Mechanical Engineering Department, Virginia Tech, Blacksburg, VA 24061, USA

*Corresponding Author Email Address: miwa@n.t.u-tokyo.ac.jp

**Abstract**

In recent years, image processing methods for gas-liquid two-phase flow, including conventional computer vision techniques, bubble detection, segmentation, and tracking algorithms, have seen significant development due to their high efficiency and accuracy. Nevertheless, obtaining extensive, high-quality two-phase flow images continues to be a time-intensive and costly process. To address this issue, a generative AI architecture called bubbly flow generative adversarial networks (BF-GAN) is developed, designed to generate realistic and high-quality bubbly flow images through physically conditioned inputs, namely superficial gas ($j_g$) and liquid ($j_f$) velocities.

Initially, 105 sets of two-phase flow experiments under varying conditions are conducted to collect 278,000 bubbly flow images with physical labels of $j_g$ and $j_f$ as training data. A multi-scale loss function of GAN is then developed, incorporating mismatch loss and feature loss to further enhance the generative performance of BF-GAN. The BF-GAN's results indicate that it has surpassed conventional GAN in all generative AI indicators, establishing for the first time a quantitative benchmark in the domain of bubbly flow. In terms of image correspondence, BF-GAN and the experimental images exhibit good agreement. Key physical parameters of bubbly flow images generated by the BF-GAN, including void fraction, aspect ratio, Sauter mean diameter, and interfacial area concentration, are extracted and compared with those from experimental images. This comparison validates the accuracy of BF-GAN's two-phase flow parameters with errors ranging between 2.3% and 16.6%. The comparative analysis demonstrates that the BF-GAN is capable of generating realistic and high-quality bubbly flow images for any given $j_g$ and $j_f$ within the research scope, and these images align with physical properties.

BF-GAN offers a generative AI solution for two-phase flow research, substantially lowering the time and cost required to obtain high-quality data. In addition, it can function as a benchmark dataset generator for bubbly flow detection and segmentation algorithms, enhancing overall productivity in this research domain. The BF-GAN model is available online (https://github.com/zhouzhouwen/BF-GAN).

# 1 Introduction

In modern engineering, solving complex physical problems typically depends on conventional simulation and experimental approaches, and it is essential to obtain extensive experimental data to progressively develop mechanistic models and empirical correlations. The emergence of artificial intelligence (AI) has introduced a new approach driven by AI, so called AI-driven methodology. This methodology uses datasets collected through conventional simulations and experiments, along with developed physical models, as training data and loss function for AI models. This has led to the development of data-driven or physics-informed AI models, such as those for parameters prediction, classification, clustering, object detection, segmentation, tracking, and physics-informed neural networks (PINNs) that incorporate empirical correlations or mechanistic models. The methodology mentioned above is depicted in **Fig. 1**.

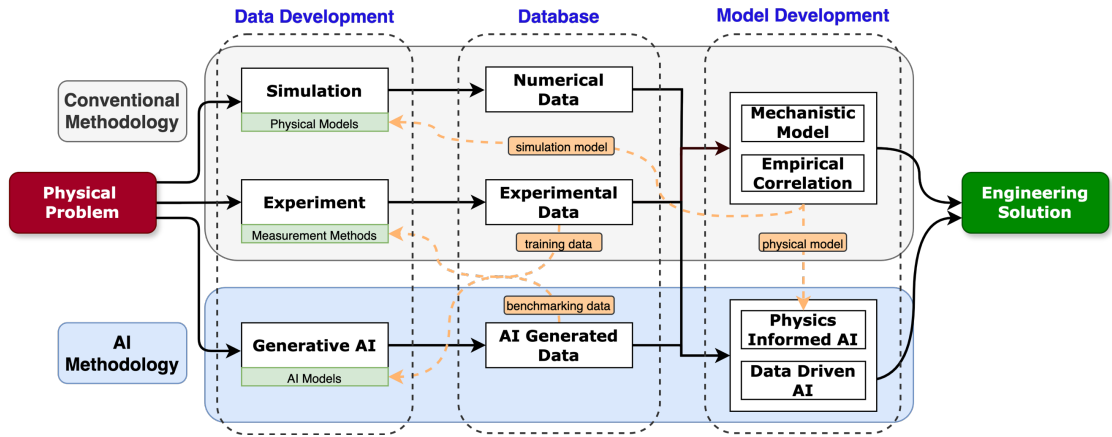

Fig. 1. General process of AI and conventional methodologies for a physical problem.

However, both conventional methods and AI-based models come with high costs and substantial time investment, especially for acquiring large volumes of high-quality data. As the complexity of the problem increases, data acquisition becomes even more challenging. Consequently, a key limitation of the AI-driven methodology is its reliance on experimental and simulation-based methods for data generation, resulting in a lack of a critical component in the data development process. In this context, the emergence of generative AI addresses this gap by providing a vital solution for AI-driven data generation, as illustrated in **Fig. 1**. By producing large amounts of high-quality synthetic data, generative AI enhances data availability, significantly accelerating the problem-solving process and providing a powerful AI-based methodology for addressing complex engineering challenges.

The integration of generative AI, with conventional engineering techniques open new possibilities for tackling multiphase flow problems. One such example is bubbly flow, which is a type of gas-liquid two-phase flow characterized by the dispersion of gas bubbles within a continuous liquid phase. The bubbly flow regime is distinguished by the presence of numerous bubbles that vary in size and distribution, moving through the liquid medium. The behavior of these bubbles, including their formation, coalescence, break-up, and interaction with the liquid, significantly influences the mass transfer, heat transfer, and mixing efficiency within various industrial applications [1-5]. In the chemical industry, the distribution and size of bubbles in catalytic reactors can affect the surface area available for reactions, thus impacting the overall chemical processes of reaction rates, heat transfer, and mass transfer [6-8]. In nuclear reactors coolant systems often involve gas-liquid mixtures, where the presence of bubbles can affect heat transfer rates and system stability [9-13]. Similarly, in the development of next-generation energy technologies, such as hydrogen production via electrolysis, the formation and detachment of gas bubbles at electrode surfaces are critical factors that determine the efficiency of the process [14-17]. Understanding the complex interactions and dynamics of bubbles within these systems is therefore essential for driving innovation and achieving optimal performance in various applications.

Image processing methods, shown in **Fig. 2**, encompass advanced AI-based bubble detection and tracking algorithms [18], segmentation techniques [19], or conventional computer vision technologies [20], have emerged as pivotal research tools for non-invasive detection of bubbly flow characteristics. These methods have significantly enhanced the efficiency and accuracy of detecting and analyzing bubbly flows.

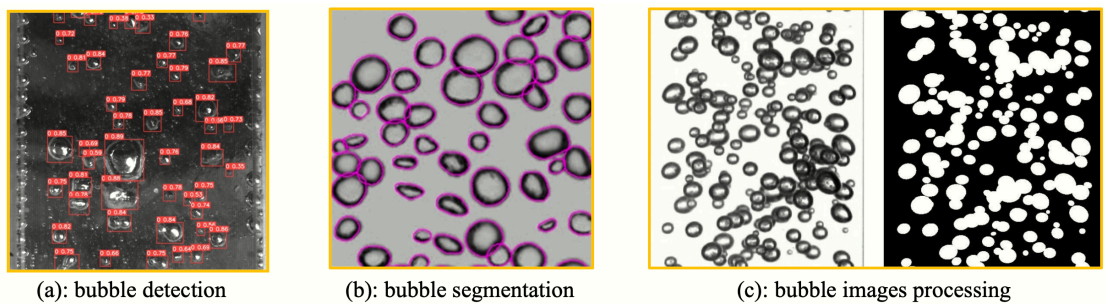

(a): bubble detection    (b): bubble segmentation    (c): bubble images processing

**Fig. 2. Image processing methods for bubbly flow.**

Despite these advancements, a major challenge remains: the need for large quantities of high-quality bubbly flow images as benchmark datasets. Capturing these images requires meticulous experimental setups, including the construction of specialized test loops and the utilization of high-speed cameras capable of recording the transient behaviors of bubbles in bubbly flows. These steps are not only time-consuming but also financially demanding, as it involves repeated experiments to ensure the accuracy and reliability of the collected data.

To further enhance the methods for obtaining large quantities of high-quality bubbly flow images, current research primarily focuses on two approaches:

The first approach utilizes conventional image processing techniques based on numerical computation, specifically the circular concentric approximation (CCA) method [21, 22]. This method assumes that bubble edge intensities follow a concentric circular/elliptical arrangement. Bubbly flow images are synthesized based on predefined bubble shapes and certain distribution information. However, the method fails to simulate the intricate structures of bubble shape and intensity variations, limiting them to generating simple bubble shapes, such as spherical or elliptical bubbles. Synthetic bubbles of different sizes may exhibit similar intensity distributions. Consequently, these synthetic images perform significantly differently when comparing with real bubbly flow images.

The second approach involves the single bubble generation model BubGAN [23]. While the generation of individual bubbles is improved, the results fail to produce realistic images of bubbly flow where multiple bubbles exist in a single image frame. Additionally, BubGAN cannot directly generate images from physical parameters such as superficial gas and liquid velocities. Generating an entire bubbly flow image requires stitching together each individual bubble, which is highly time-consuming. Although this method effectively enhances the accuracy of local bubble morphology, it is limited in capturing global features.

The significant advancements in text-to-image generative AI have provided valuable solutions for generating high-fidelity bubbly flow images. Prominent text-to-image AI models like DALL-E 2 [24], Stable Diffusion [25], and MidJourney [26] have made significant strides in generating high-fidelity images from textual descriptions. **Fig. 3** is generated from the text prompt: "A student in a University of Tokyo classroom is reading Energy." However, while these models are capable of generating a wide range of images and artwork, they have certain limitations when it comes to producing bubbly flow images. First, the interpretation of text prompts can be inconsistent, which may result in variations in image details such as lighting, contrast, and texture. Second, precise control over bubble size, shape, and location can be very challenging with these models. Achieving the desired images may require highly detailed and specific prompts. Additionally, the process of translating text into images is not fully transparent, which makes it difficult to interpret two-phase flow behavior in the generated images.

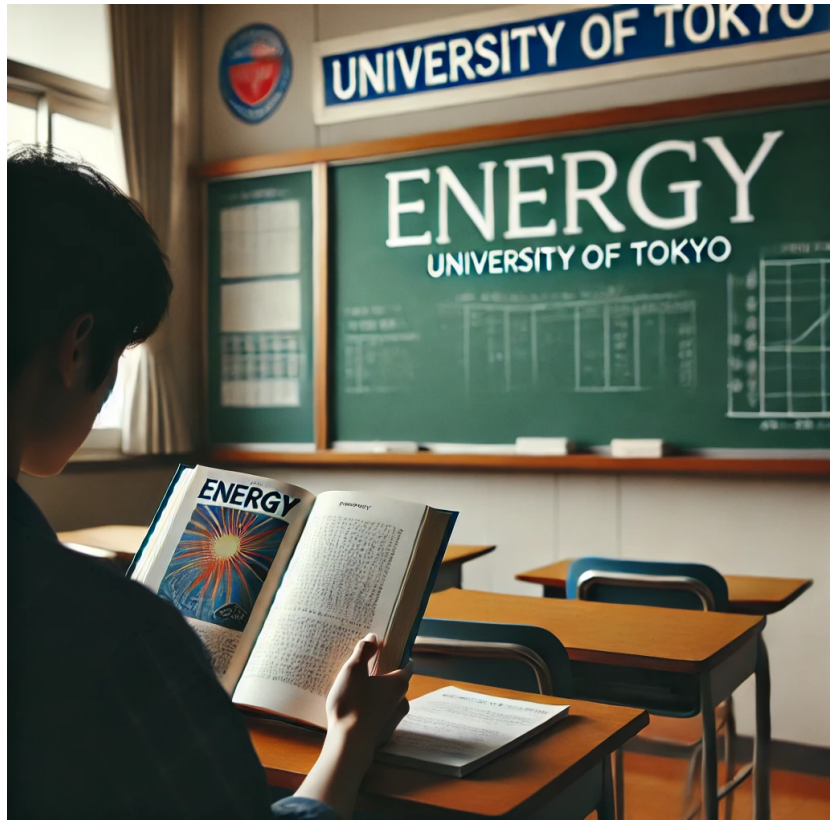

Fig. 3. Images generated by the open-source model DALL-E 2.

In the present study, a generative AI architecture termed Bubbly Flow Generative Adversarial Networks (BF-GAN) is developed, which is designed to generate realistic and high-quality bubbly flow images from physically conditioned inputs, namely, superficial gas velocity $j_g$ and superficial liquid velocity $j_f$. To train the BF-GAN, 105 sets of bubbly flow experiments with varying $j_g$ and $j_f$ are conducted, resulting in a dataset of 278,000 images. A generator developed by NVIDIA is employed to learn the features of bubbly flows. Additionally, a multi-scale loss, encompassing mismatch loss and feature loss, is incorporated into the BF-GAN to further enhance its generative performance. The generative capability of the BF-GAN is subsequently validated, demonstrating comprehensive superiority over conventional GANs in AI indicator and providing, for the first time, quantifiable benchmarks in the domain of bubbly flow generative AI. Regarding the image correspondence, BF-GAN and the experimental images exhibit good agreement. In terms of physical indicators, void fraction, bubble aspect ratio, Sauter mean diameter, and interfacial area concentration are extracted and compared with experimental images, further validating the BF-GAN's physical performance. The BF-GAN is open-sourced and available in the GitHub repository, accompanied by detailed installation and usage instructions. (https://github.com/zhouzhouwen/BF-GAN)

## 2 Method
### 2.1 Experimental setup
The experimental setup consists of a vertical upward two-phase flow loop, depicted in **Fig. 4**, designed to operate at room temperature and atmospheric pressure. Water is drawn from a 1 m³ tank by a centrifugal pump, and its flow rate is recorded using a magnetic flowmeter. Air is supplied from a buffer tank, maintained at a pressure of 0.7 MPa by an air compressor, with its mass flow rate measured by airflow sensors, pressure transducers, and a K-type thermocouple. The flow rates of both air and water are regulated by control valves and introduced into the test section via a two-phase mixer made of polyvinyl chloride and porous materials. This mixture travels through a section of clear acrylic pipe (25.4 mm internal diameter and 1 m length), enabling the observation of flow regimes. High-speed imaging is conducted 140 cm above the mixer outlet, corresponding to a length-to-diameter ratio of 55.1. After passing through the test section, the air-water mixture returns to the water tank, where it is naturally separated by gravity.

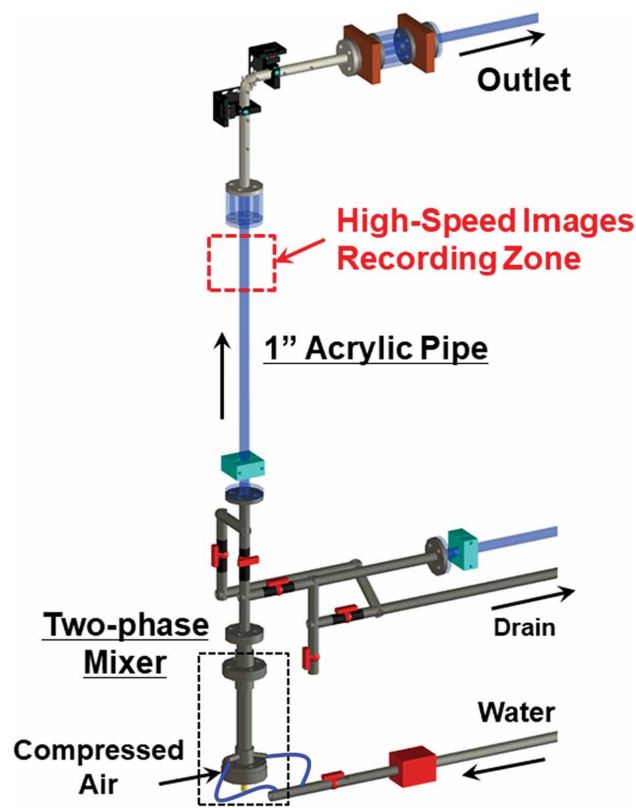

**Fig. 4. Two-phase flow test loop.**

**2.2 BF-GAN**

Conditional generative adversarial network (CGAN) is a class of generative AI deep learning frameworks [27], which is designed to generate new data samples that are indistinguishable from real data. CGAN consist of two main components: the generator and the discriminator, which engage in a two-player-zero-sum game for Nash equilibrium.

The generator is a neural network tasked with producing data samples that resemble the training data. It takes random noise and feature matrix as input and transforms it into a data sample through a series of nonlinear transformations. Typically, the generator is composed of a deep neural network, such as a feedforward network or a convolutional neural network (CNN) [28]. The primary goal of the generator is to cheat the discriminator into believing that the generated data is real. A schematic diagram of a generator network is shown in **Fig. 5**. This generator follows an encoder-decoder framework, utilizing convolutional and deconvolutional layers to effectively process and transform the input into realistic images. The encoding begins with a convolutional layer that transforms the input into a $512 \times 512 \times 64$ tensor. Subsequent layers further downsample the spatial dimensions while increasing the depth of the feature maps. The architecture employs Leaky ReLU (LReLU) activations and Batch Normalization (BN) after each convolutional layer to enhance the learning stability and convergence. The final encoding is achieved through a series of $1 \times 1$ convolutions, which compress the information into a $1 \times 1 \times n$ vector, where $n$ represents the latent space dimensions. The decoding phase involves upsampling the encoded representation back to the original image dimensions through deconvolutional layers. This process is designed to reconstruct high-resolution images that retain the characteristics defined by the input feature matrix.

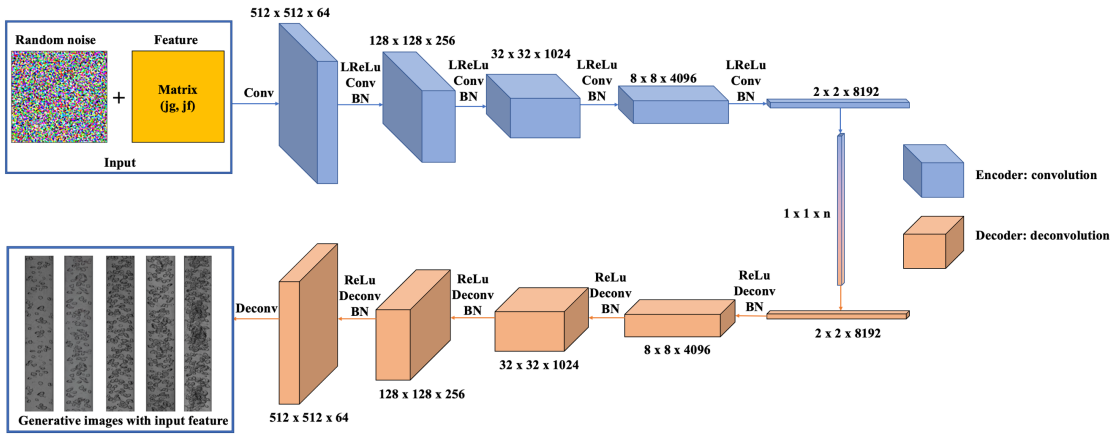

**Fig. 5. A schematic diagram of a generator network of a CGAN.**

The discriminator, on the other hand, is a neural network designed to distinguish between real data and fake data generated by the generator. It is often implemented as a deep neural network, specifically a CNN due to its efficacy in handling image data. The discriminator receives both real and fake data samples and outputs a probability indicating whether the sample is real or fake (generated). The objective of the discriminator is to correctly classify the input data, thereby improving its ability to detect fake samples. A schematic diagram of a generator network is shown in **Fig. 6**. The discriminator employs several convolutional layers, each followed by activation functions and normalization techniques, to extract and process the features from the input images. The convolutional layers progressively reduce the spatial dimensions of the input while increasing the depth of the feature maps, enabling the network to capture features. The final layer of the discriminator is a convolutional layer with a Sigmoid activation function, which outputs a single-channel image with dimensions $512 \times 512$

× 1. The Sigmoid function is used to produce a probability map, where each pixel value ranges between 0 and 1, indicating the likelihood of the corresponding input region being real or fake.

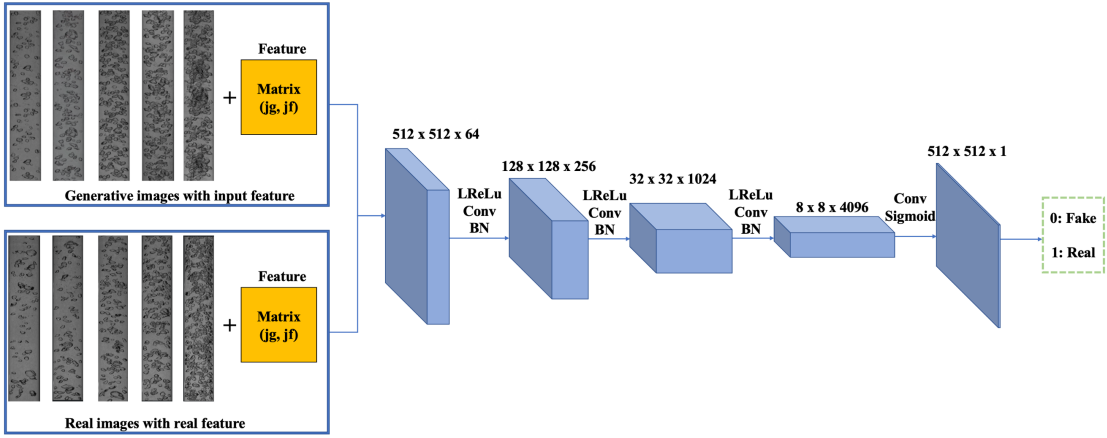

**Fig. 6. A schematic diagram of a discriminator network of a CGAN.**

The training of CGAN involves a simultaneous optimization process where both networks are trained together in a zero-sum game. The generator aims to minimize the following loss function:

$$Loss_G = -\mathbb{E}_{z \sim p_z(z)}[\log(D(G(z,c)))] \tag{1}$$

where $G(z,c)$ is the output of the generator given input noise $z$ and feature matrix $c$, and $D(G(z,c))$ is the probability assigned by the discriminator to the generated sample being real.

Conversely, the discriminator aims to maximize its classification accuracy using the following loss function:

$$Loss_D = -\mathbb{E}_{x \sim p_{data}(x)}[\log(D(x,c))] - \mathbb{E}_{z \sim p_z(z)}[\log(1 - D(G(z,c)))] \tag{2}$$

here, $D(x)$ represents the probability that the discriminator assigns to a real data sample $x$ with corresponding feature matrix $c$ being real, and $D(G(z,c))$ is the probability assigned to a generated sample being real.

The overall objective of the CGAN can be expressed as a minimax optimization problem:

$$\min_G \max_D Loss(D,G) = \mathbb{E}_{x \sim p_{data}(x)}[\log D(x,c)] + \mathbb{E}_{z \sim p_z(z)}[\log(1 - D(G(z,c)))] \tag{3}$$

This formulation illustrates the adversarial nature of CGAN, where the generator and discriminator are in constant competition. As training progresses, the generator becomes increasingly adept at producing realistic data, while the discriminator becomes better at identifying generated data. The objective of training is to develop an effective generator network for future image generation. Essentially, generative AI models are probabilistic models in a multidimensional space. **Fig. 7** visualizes the training process of a CGAN model, depicted as a blue point cloud in three-dimensional space. The red points represent a true distribution with a mean of [0, 0, 0] and a covariance of [1, 1, 1]. As training progresses, the probability distribution of the CGAN gradually approximates the real distribution.

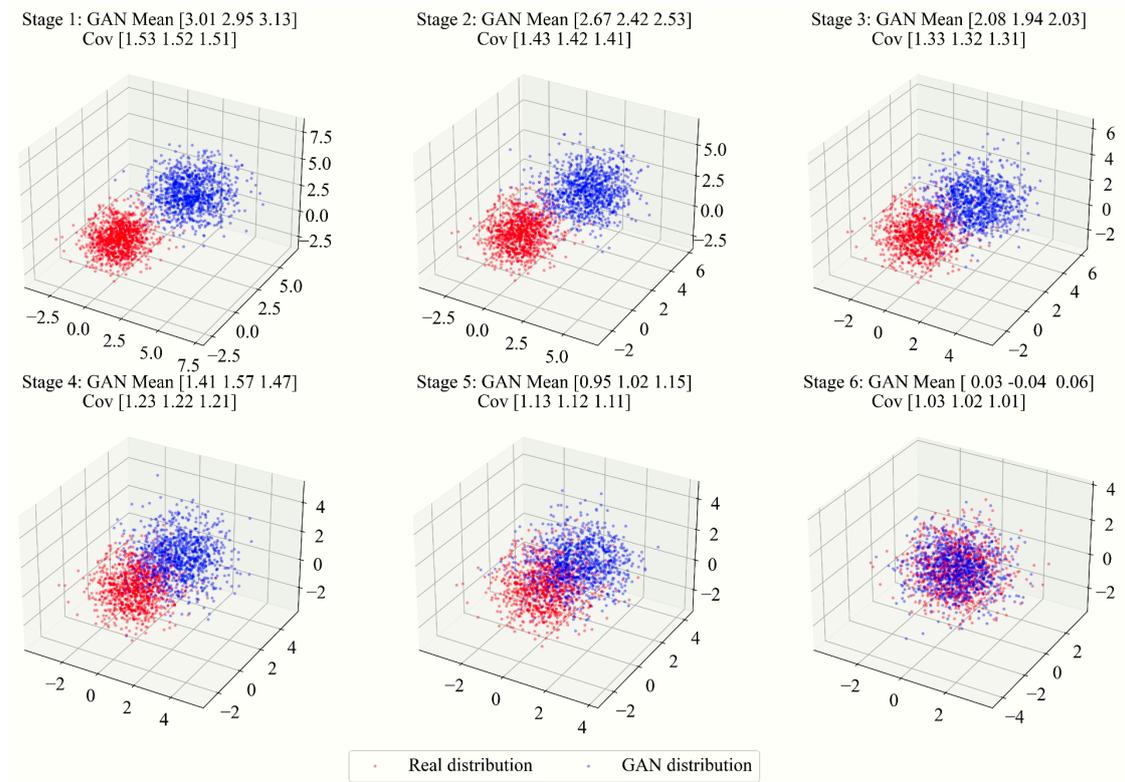

**Fig. 7. Visualization of the generative AI model training process.**

Conventional CGAN face several significant challenges, particularly related to the effectiveness of their generators [29] and the inherent limitations of the original CGAN loss functions [30]. One major issue is the generator's difficulty in producing high-fidelity, realistic images, often resulting in outputs that lack sharpness and exhibit noticeable artifacts. This problem is partly due to the instability commonly observed during training, where the generator learns to produce a limited variety of outputs rather than capturing the full diversity of the data distribution. Furthermore, the original GAN loss function, especially when the discriminator with conditional input becomes too strong, provides little useful feedback to the generator. This imbalance disrupts the learning process of generator.

To address these limitations and further enhance the generative performance of BF-

GAN, the generator architecture developed by NVIDIA [31], as depicted in **Fig. 8**, has been utilized. This generator transforms the noise into an intermediate latent space $W$, which enables more stable and controllable manipulations of the generated images. The generator also employs Fourier features and 1×1 convolution layers to facilitate better integration of the input latent space and spatial features, enhancing the quality of the generated images. Each layer in the network, denoted as L0-L13, is designed to progressively refine the image resolution, ensuring fine-grained details are captured accurately. The incorporation of exponential moving averages (EMA) in the weight updates further stabilizes the training process by smoothing out the parameter updates. Additionally, the custom CUDA kernel enhances computational efficiency, enabling the generator to perform complex transformations such as upsampling, downsampling, and cropping with high performance.

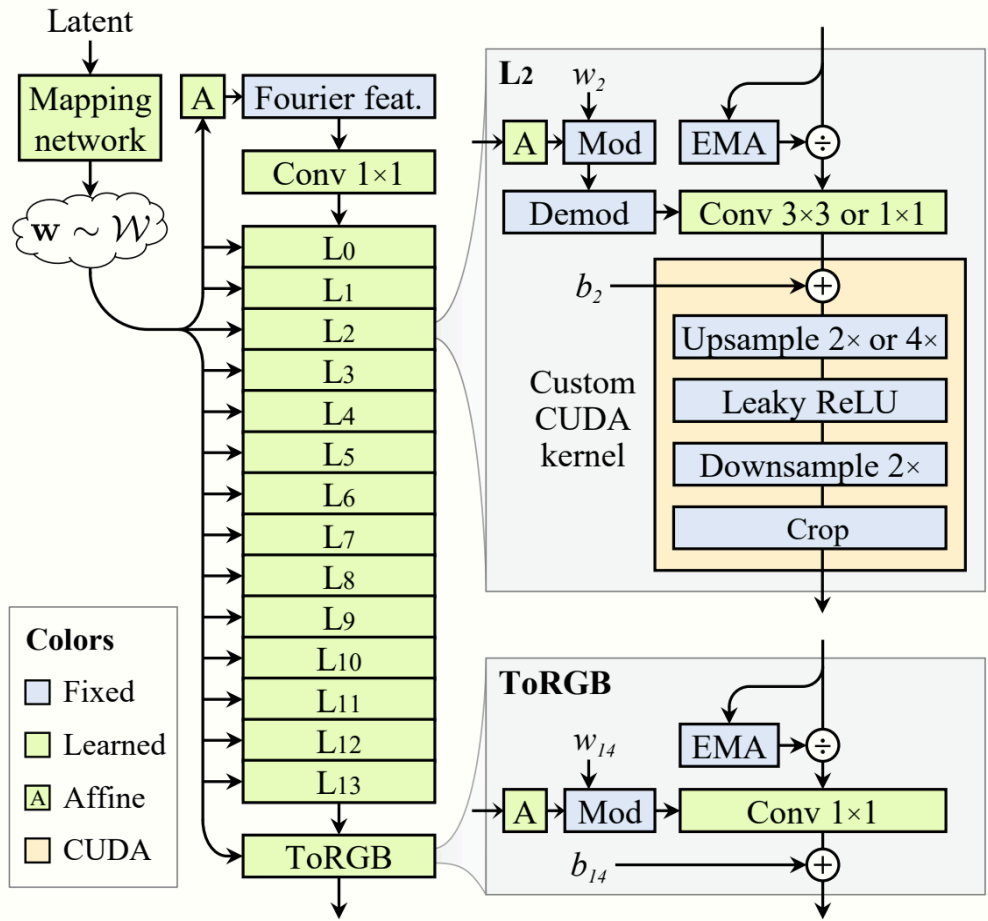

**Fig. 8. Generator architecture developed by NVIDIA.**

Moreover, a multi-scale loss has been developed to improve the generative performance. This includes the incorporation of mismatch loss, which distinguishes between correct and incorrect matching conditions, thereby ensuring that the generator produces images that are not only realistic but also contextually accurate according to the given conditions:

$$\min_G \max_D Loss(D,G) = \mathbb{E}_{x \sim p_{\text{data}}(x)}[\log D(x, c_t) + \log(1 - D(x, c_f))] +$$
$$\mathbb{E}_{z \sim p_z(z)}[\log(1 - D(G(z, c_t))) + \log(1 - D(G(z, c_f)))] \qquad (4)$$

Here, $c_t$ indicates the true condition and $c_f$ represents the false condition, $c_f$ are randomly generated during the training process.

Additionally, feature loss has been integrated, leveraging the L1 and L2 distances to penalize average discrepancies in the features extracted by a pre-trained VGG network [32]. The VGG network is a convolutional neural network architecture recognized for its deep structure of up to 19 layers. It employs sequences of convolutional layers with small receptive fields of 3x3, followed by max pooling layers, which collectively enable the network to effectively capture complex features at multiple scales. This approach ensures that the differences are measured in a more meaningful feature space, capturing perceptual discrepancies that are more aligned with human visual perception. The L1 and L2 loss are defined as:

$$L1(G) = \mathbb{E}_{x \sim p_{\text{data}}(x), z \sim p_z(z)}[||F(x) - F(G(z, c_t))||_1] \qquad (5)$$

$$L2(G) = \mathbb{E}_{x \sim p_{\text{data}}(x), z \sim p_z(z)}[||F(x) - F(G(z, c_t))||_2^2] \qquad (6)$$

where $F(\cdot)$ represents the feature extraction function of the pre-trained VGG network. The schematic diagram of the multi-scale loss function is shown in **Fig. 9**.

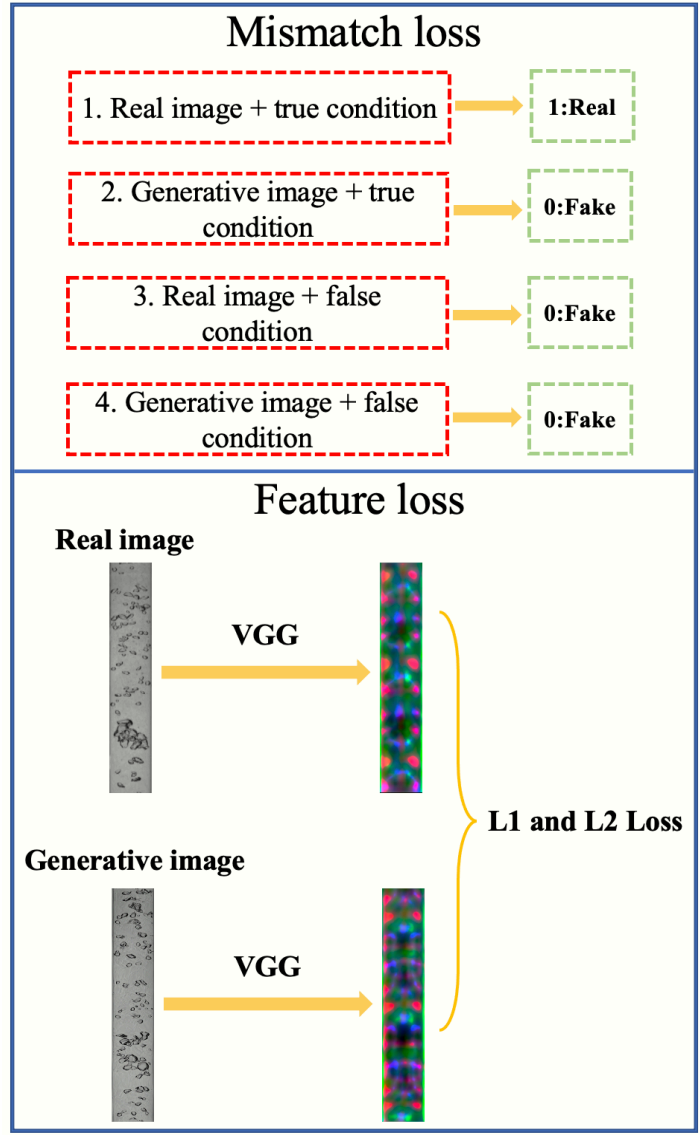

Fig. 9. Schematic diagram of the multi-scale loss function.

In summary, the algorithm flow of the BF-GAN is shown in **Algorithm 1**.

**Algorithm 1. Developed BF-GAN in the present study.**

**Input:** $random\ noise, j_g$ and $j_f$
**Output:** $image$

**1. Initialization:**
    (1) Initialize the generator $G$ and discriminator $D$ with random weights.
    (2) Prepare real data samples $x$ and corresponding conditions $c_t$ (true conditions).
    (3) Generate random noise $z$.
    (4) Randomly generate false conditions $c_f$ for discriminator training.
    (5) Epoch = 0.

**2. Training loop: while Epoch < Epoch$_{max}$ do:**
    **2.1 Discriminator update:**

(1) Sample a batch of real data samples $x$ and true conditions $c_t$.
(2) Generate a batch of fake data samples using the generator: $\tilde{x} = G(z, c_t)$.
(3) Compute the discriminator loss for real and fake data:

$$Loss_D = -\mathbb{E}_{x \sim p_{\text{data}}(x)}\left[\log D(x, c_t) + \log\left(1 - D(x, c_f)\right)\right]$$

$$- \mathbb{E}_{z \sim p_z(z)}[\log\left(1 - D(G(z, c_t))\right) + \log\left(1 - D(G(z, c_f))\right)]$$

(4) Update the discriminator parameters via gradient descent.

**2.2 Generator update:**

(1) Generate a batch of fake data samples using the generator: $\tilde{x} = G(z, c_t)$.
(2) Compute the feature loss using a pre-trained VGG network:

$$\text{L1}(G) = \mathbb{E}_{x \sim p_{\text{data}}(x), z \sim p_z(z)}[||F(x) - F(G(z, c_t))||_1]$$

$$\text{L2}(G) = \mathbb{E}_{x \sim p_{\text{data}}(x), z \sim p_z(z)}[||F(x) - F(G(z, c_t))||_2^2]$$

$$Loss_{feature} = \text{L1}(G) + \text{L2}(G)$$

(3) Compute the GAN loss for the generator:

$$Loss_G = -\mathbb{E}_{z \sim p_z(z)}[\log(D(G(z, c_t)))]$$

(4) Combine the GAN loss and feature loss:

$$Total\ loss(G) = Loss_{feature} + Loss_G$$

(5) Update the generator parameters via gradient descent.

**end**

**3. Output:**

(1) Once the training converges, the generator $G$ is capable of producing high-quality images conditioned on the input features $c_t$.

## 2.3 Bubble detection model based on YOLO

To quantify and validate the physical properties of images generated by the BF-GAN, a bubble detection model based on You Only Look Once (YOLO) has been developed in the previous study [33].

YOLO is a state-of-the-art, real-time object detection AI framework designed for speed and accuracy in detecting objects within an image [34]. It predicts the locations and categories of objects through a single forward pass of the network, significantly improving processing speed. Unlike conventional object detection methods, YOLO frames detection as a single regression problem, directly mapping from image pixels to bounding box coordinates and class probabilities. This integrated approach allows YOLO to achieve high bubble detection accuracies while maintaining real-time processing speeds.

In previous research, approximately 600 bubbles were manually annotated using the Labelme software to create a training dataset. By training with YOLO, a bubble detection model based on YOLO has been developed and validated. The performance of bubble detection is shown in **Fig. 10**.

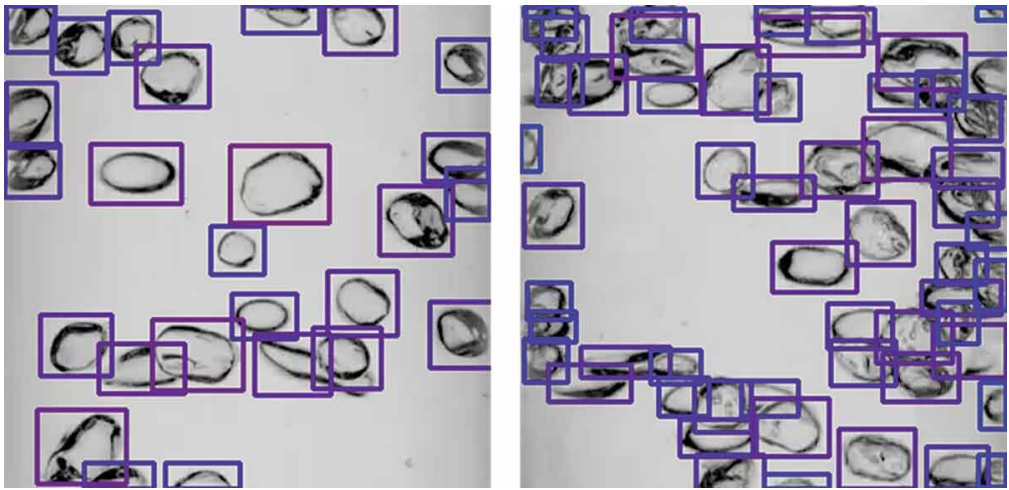

**Fig. 10. Bubble detection results by the YOLO model.**

Using this bubble detection model, each bubble in a bubbly flow image can be detected well. The bounding box coordinates of the detected bubbles are extracted and converted to real-world dimensions. This enables the extraction of four key bubbly flow parameters from a bubbly image: void fraction, aspect ratio, Sauter mean diameter, and interfacial area concentration.

The development workflow of the BF-GAN is illustrated in **Fig. 11.**
**Step 1:** Initially, the research flow pattern was identified as the bubbly flow region within the Mishima-Ishii flow regime map.
**Step 2:** Videos of the bubbly flow under each condition were recorded and segmented into individual frames.
**Step 3:** In the dataset, labels were assigned to all images under each specific $j_g$ and $j_f$ condition, and each was assigned a unique address.
**Step 4:** The BF-GAN was trained using the prepared dataset.
**Step 5:** Manual optimization of the BF-GAN parameters was performed to achieve the optimal model. Upon inputting the $j_g$ and $j_f$ conditions, the BF-GAN generates the corresponding bubbly flow images.
**Step 6:** The authenticity of the images generated by the BF-GAN was verified through AI indicator, image correspondence, and two-phase flow parameters.

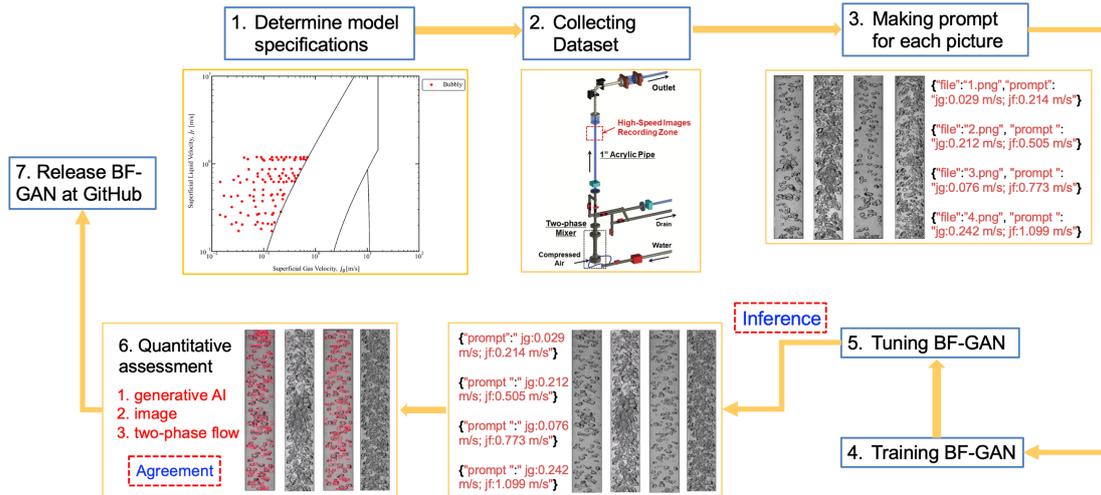

**Fig. 11. Development workflow of the BF-GAN.**

## 3 Results and discussions
### 3.1 Collection of datasets

In the present study, 105 sets of experiments under varying $j_g$ and $j_f$ conditions were conducted within the bubbly flow region, as categorized according to the Mishima-Ishii flow regime map and depicted in **Fig. 12**. Each experimental session was recorded for 200 seconds at a rate of either 10 or 20 frames per second, resulting in a total of 278,000 different images of bubbly flow. The original resolution of these images was 968 x 968 pixels. Considering the training duration for the generative AI model, BF-GAN, images resized to 512 x 512 pixels were selected for the training dataset. This resolution represents a balance of efficiency for the current study, as images at 1024 pixels would entail approximately three to four times the amount of training data compared to 512 pixels. Specific training durations and configurations will be discussed in Section 3.2.

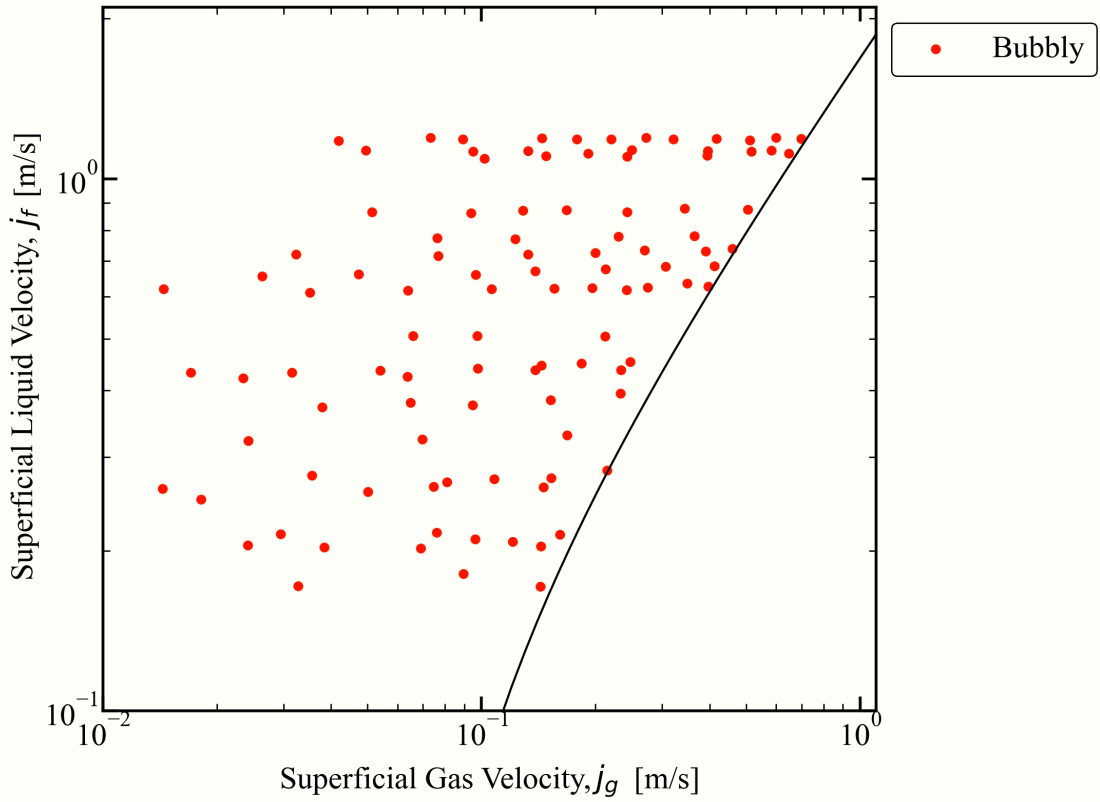

Fig. 12. Experimental conditions and dataset in the present study.

**3.2 Quantitative assessment of BF-GAN from the perspective of generative AI**

The workstations used in this study were based on Ubuntu 22.04, equipped with an NVIDIA RTX A6000 ADA GPU featuring 48 GB of memory and an Intel i9-13900 CPU with 32 cores. **Table 1** provides a detailed configuration of the BF-GAN model, including specifics of each layer and training parameters. GPU memory consumption during training ranged from approximately 30-40 GB. The training period extended roughly 12 to 13 days for each training. Including the time for parameter tuning, the entire training phase lasted about 100 days. For inference, despite the extensive training period, the GPU consumption was markedly reduced to approximately 2-3 GB, with each bubbly flow image being processed in less than 0.1 second.

**Table 1. Configuration of the BF-GAN.**

| Parameter | Value |
| --- | --- |
| Input image size | [3, 512, 512] |
| Input random noise and size | Gaussian noise, [512] |
| Condition vector size | [2] |
| Generator input size | [512] |
| Generator hidden layers | 33 generator layers |
| Generator output size | [3, 512, 512] |
| Generator total parameters | 25,137,199 |

| Parameter | Value |
| --- | --- |
| Discriminator input size | Mapping generator output and condition, [3, 512, 512] |
| Discriminator hidden layers | 42 discriminator layers |
| Discriminator output | 0 or 1 (Real or fake) |
| Discriminator total parameters | 31,347,776 |
| Optimizer | AdamW |
| Generator learning rate | 0.0025 (betas1=0, betas2=0.99) |
| Discriminator learning rate | 0.002 (betas1=0, betas2=0.99) |
| Loss function | Multi-scale loss function |
| Training epochs | 10,000 |
| Batch size | 32 |

**Fig. 13** illustrates a series of images under different $j_g$ and $j_f$ conditions, arranged from left to right: experimental images, images generated by BF-GAN, and images generated by conventional GAN. Three random frames are displayed for each type. Notably, even with the same input conditions of $j_g$ and $j_f$, BF-GAN generates varying results due to different random seeds employed. The results displayed in **Fig. 13** reveal that BF-GAN's images are highly consistent with the experimental ones, exhibiting significant realism and diversity. On the other hand, the conventional GAN generates lower-quality images with limited diversity, where even varied seeds result in similar images, maintaining consistent bubble positions and shapes across different random seeds. Additionally, these images often display noticeable artifacts, including unnatural textures, distorted edges, and inconsistent details, particularly when bubbles are close to each other.

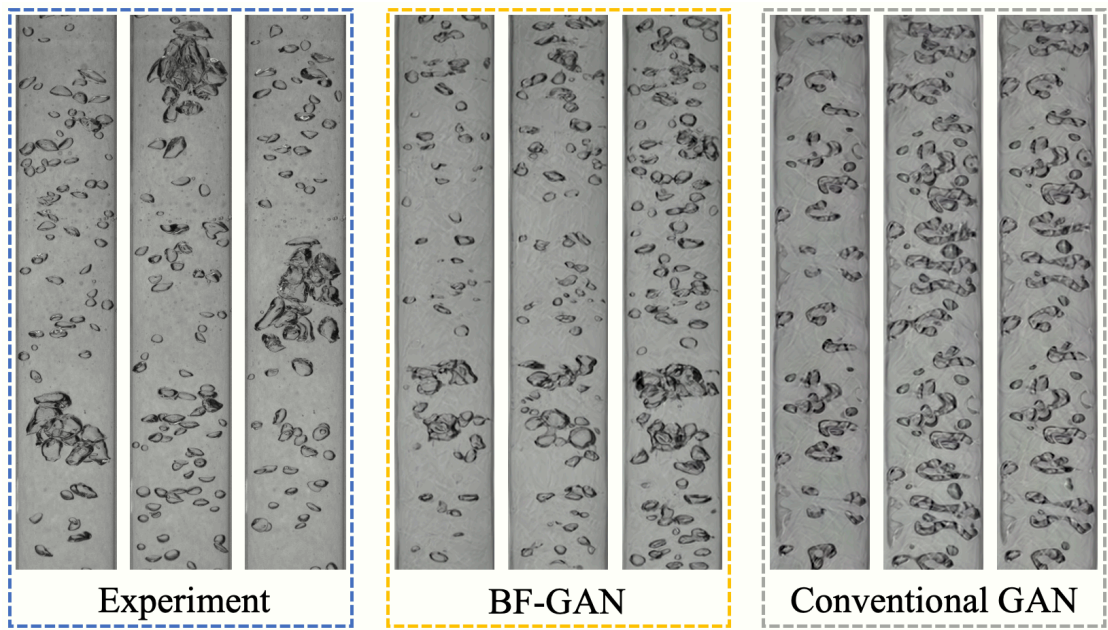

Experiment     BF-GAN     Conventional GAN

**(a)** $j_g$:0.018 *m/s*, $j_f$:0.249 *m/s*

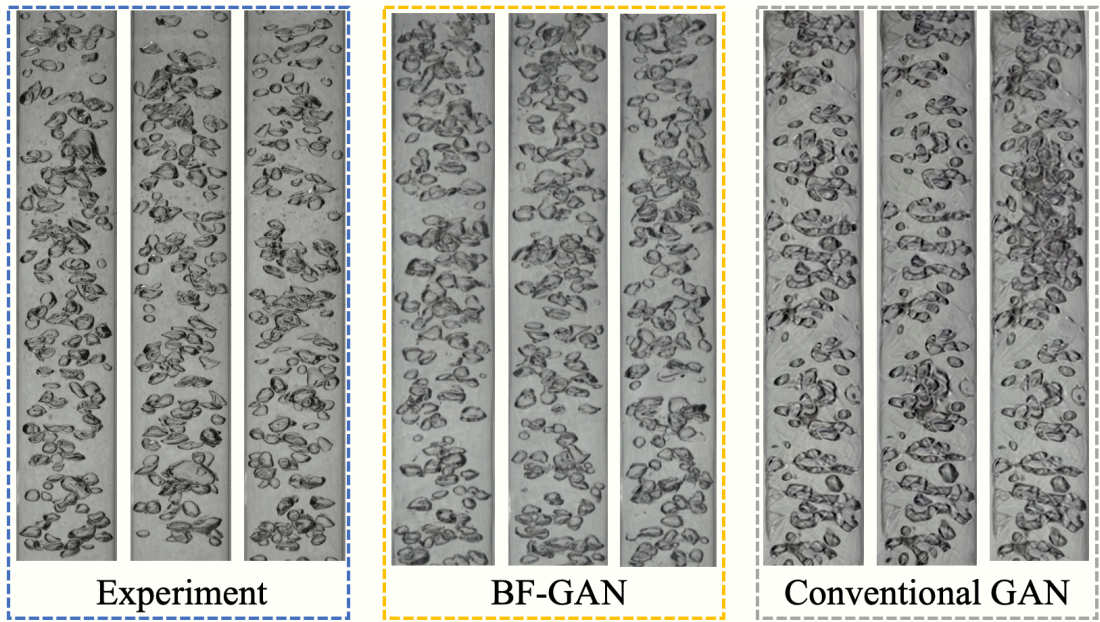

**(b)** $j_g$:0.050 *m/s*, $j_f$:0.258 *m/s*

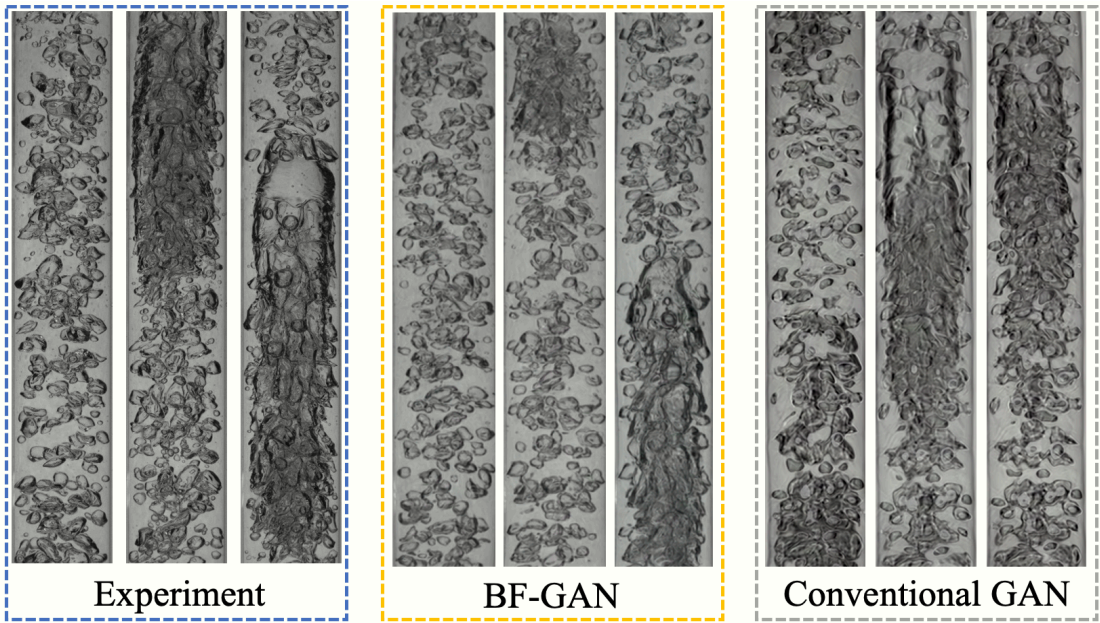

**(c)** $j_g$:0.120 *m/s*, $j_f$:0.207 *m/s*

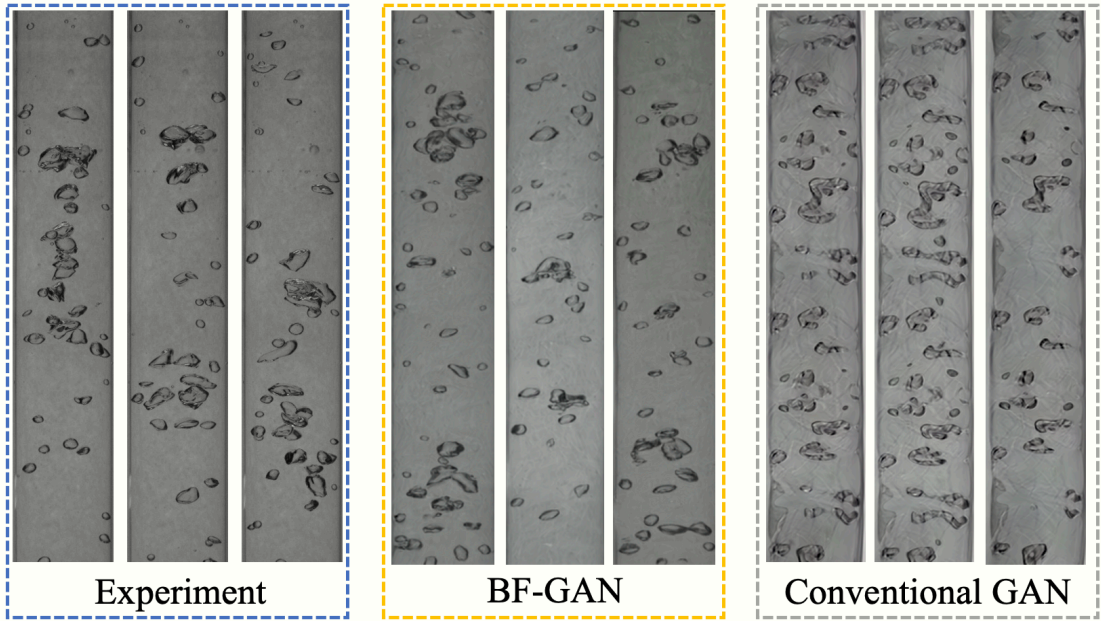

**(d)** $j_g$:0.023 *m/s*, $j_f$:0.422 *m/s*

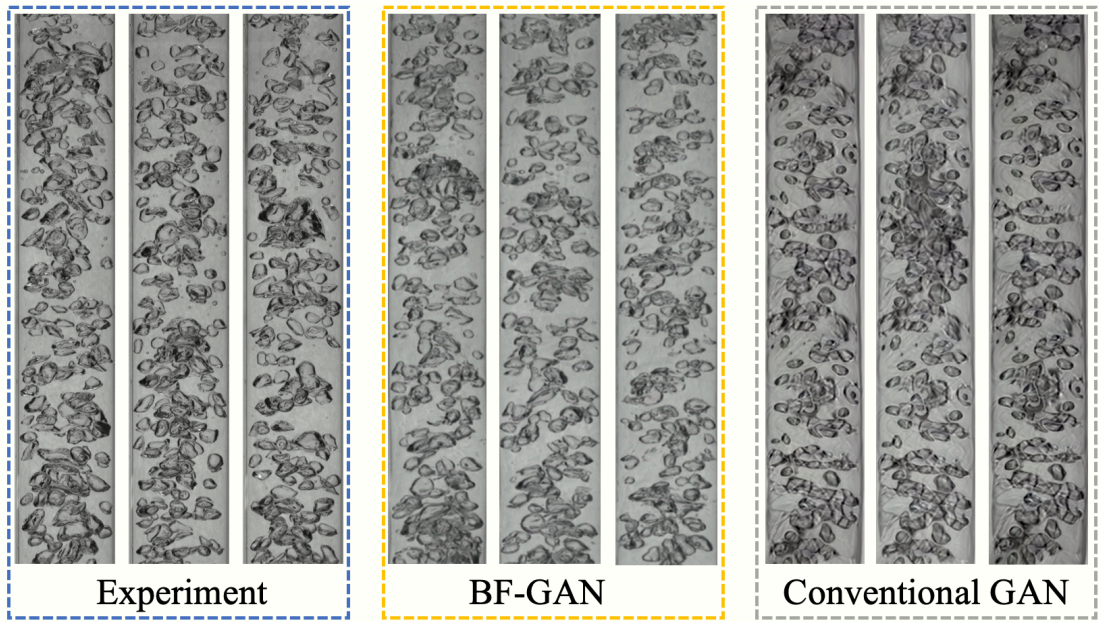

**(e)** $j_g$:0.097 *m/s*, $j_f$:0.439 *m/s*

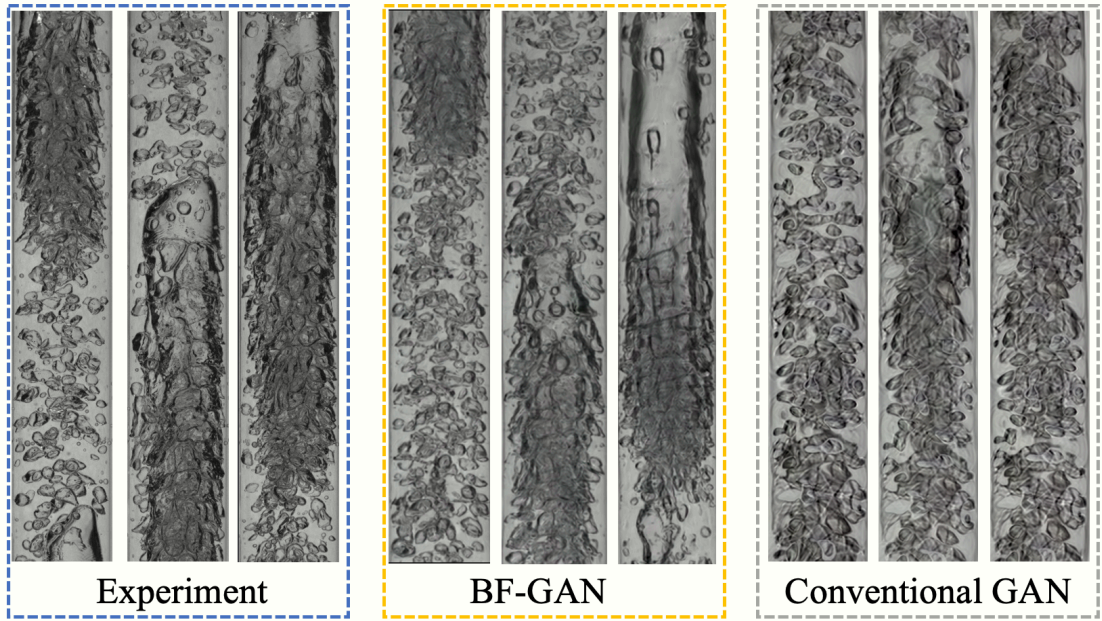

**(f)** $j_g$:0.233 *m/s*, $j_f$:0.395 *m/s*

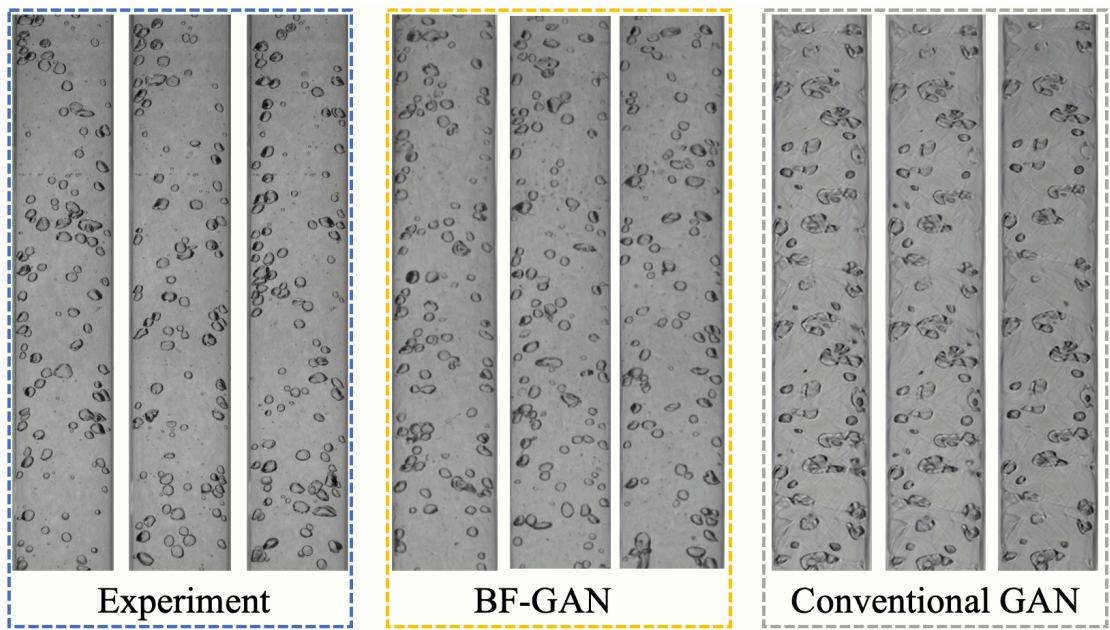

**(g)** $j_g$:0.049 *m/s*, $j_f$:1.129 *m/s*

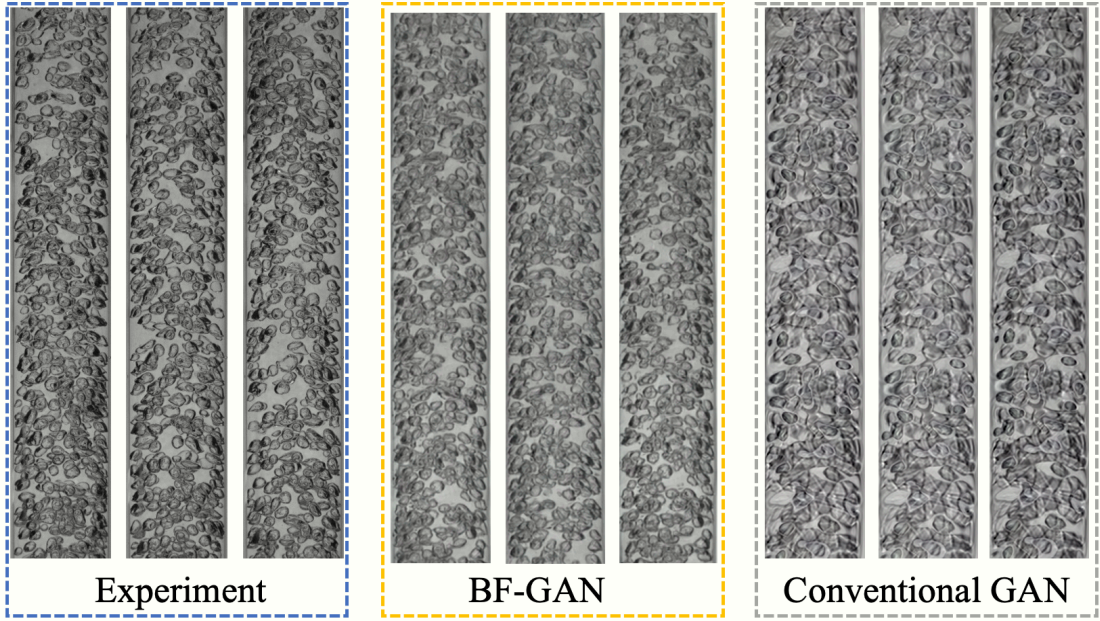

**(h)** $j_g$:0.249 *m/s*, $j_f$:1.133 *m/s*

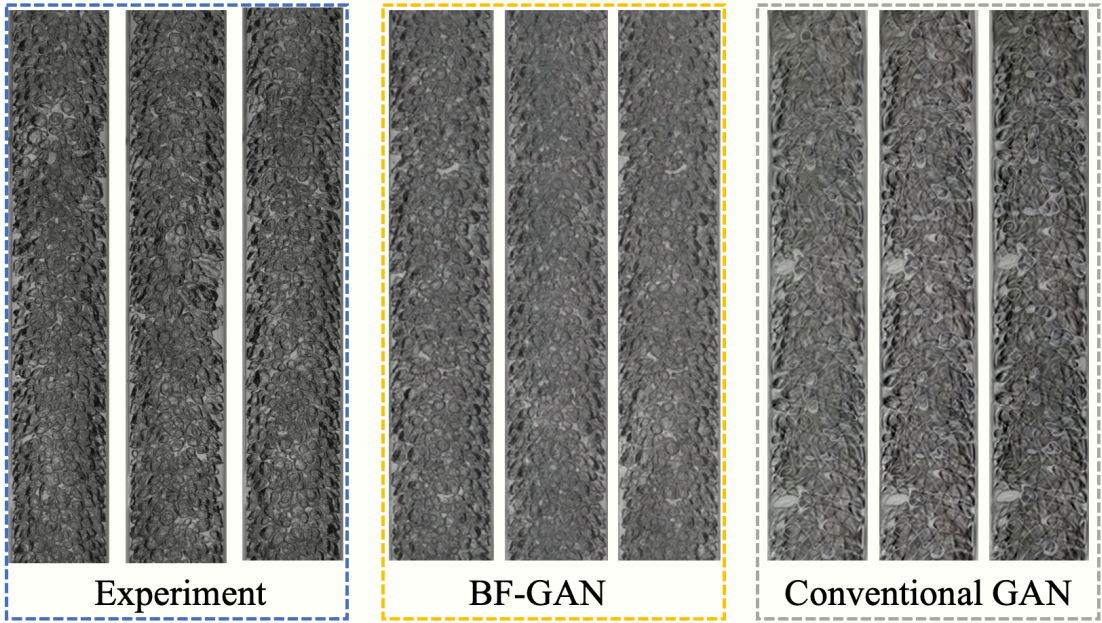

**(i)** $j_g$:0.582 *m/s*, $j_f$:1.131 *m/s*

**Fig. 13. Image comparison of experiment, BF-GAN, and conventional GAN. (a) $j_g$:0.018 *m/s*, $j_f$:0.249 *m/s*, (b) $j_g$:0.050 *m/s*, $j_f$:0.258 *m/s*, (c) $j_g$:0.120 *m/s*, $j_f$:0.207 *m/s*, (d) $j_g$:0.023 *m/s*, $j_f$:0.422 *m/s*, (e) $j_g$:0.097 *m/s*, $j_f$:0.439 *m/s*, (f) $j_g$:0.233 *m/s*, $j_f$:0.395 *m/s*, (g) $j_g$:0.049 *m/s*, $j_f$:1.129 *m/s*, (h) $j_g$:0.249 *m/s*,**

$j_f$:1.133 *m/s*, (i) $j_g$:0.582 *m/s*, $j_f$:1.131 *m/s*,

Sixteen conditions of $j_g$ and $j_f$ were uniformly selected within the bubbly flow region. **Fig. 14 (a)** illustrates the global generative performance of BF-GAN, while **Fig. 14 (b)** marks these sixteen conditions with green stars.

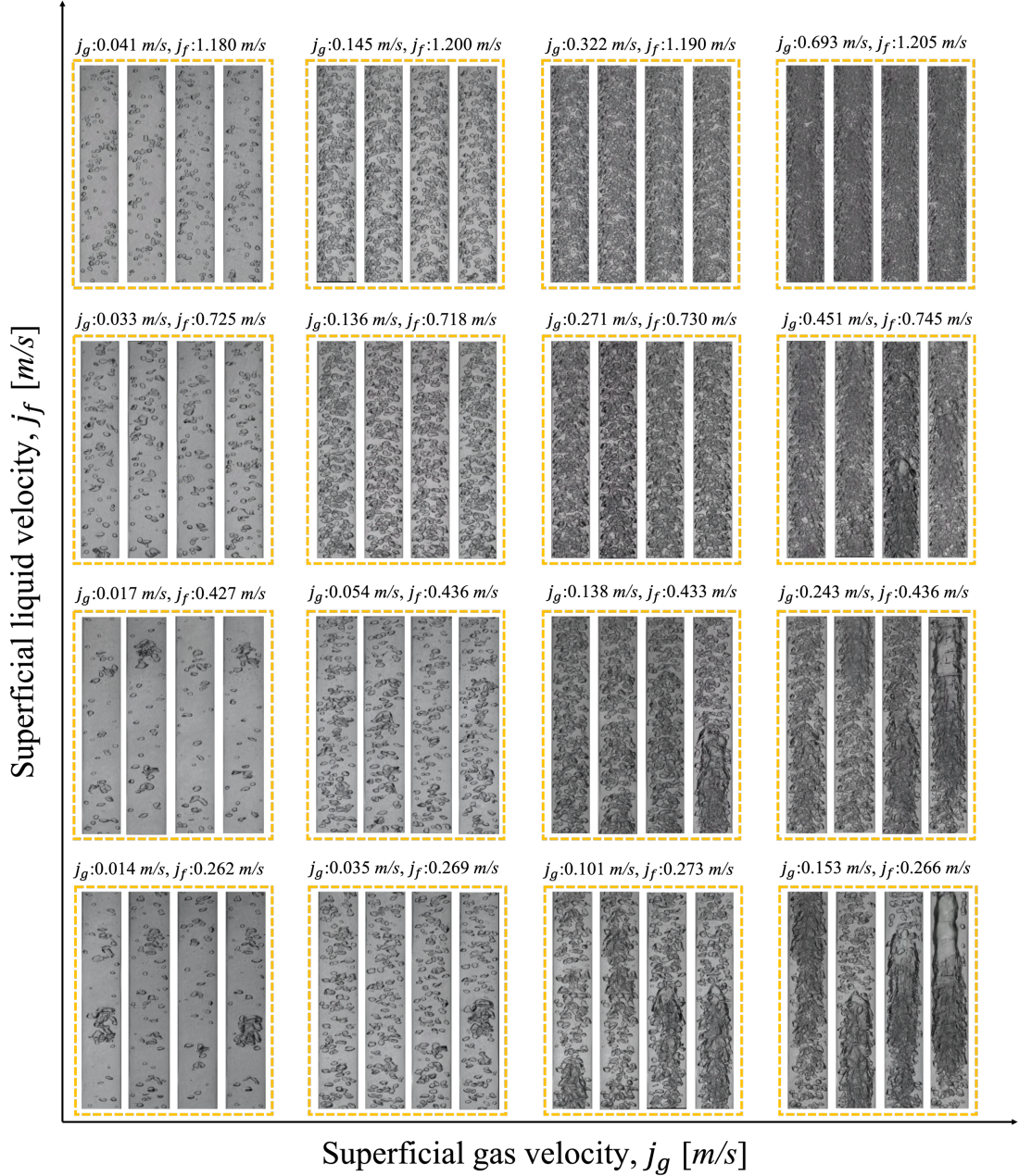

**(a) Sampling bubbly flow images according to green star conditions.**

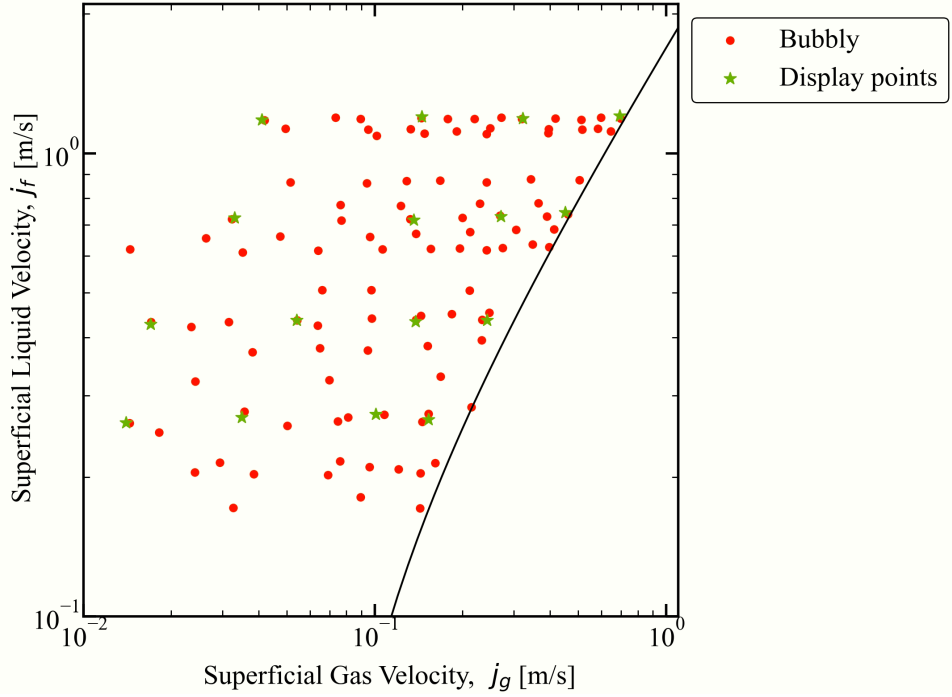

**(b) Conditions of sixteen bubbly flow images.**

**Fig. 14. Sixteen bubbly flow image samples selected based on the MI map. Green stars indicate display points.**

To quantify the generative performance of BF-GAN compared to conventional GAN, the following AI indicators were used: (1) Fréchet Inception Distance (FID), FID measures the similarity in the feature space between real and generated images. It employs the Inception model [35] to extract features and then calculates the Fréchet distance. A lower FID indicates higher image quality and greater resemblance to real images; (2) Kernel Inception Distance (KID), KID quantifies the difference between generated and real images by calculating the maximum mean discrepancy of Inception network features. Similar to FID, a lower KID is preferable; (3) Precision and Recall [36], these indicators assess the diversity and authenticity of generated images. High precision indicates a greater number of generated images resembling the real dataset, while high recall suggests the generated images capture the diversity of the real dataset. Ideally, both indicators should be large; (4) Perceptual Path Length (PPL) [37], PPL measures the visual impact of small step changes in the latent space on generated images. A lower PPL indicates smoother transitions in generating continuous images; (5) Equivariance Translation and Rotation (EQ-T, EQ-R), these indicators measure the model's invariance to translation and rotation. Ideally, if the input image is translated or rotated, the generated image should exhibit a similar change, with higher invariance being preferable; (6) Inception Score (IS) [38], IS is utilized to evaluate the quality and diversity of generated images. Higher IS values typically indicate higher quality and diversity of the images. **Table 2** shows the comparison results of AI indicators. It can be seen that BF-GAN comprehensively surpasses conventional GAN, which shows the high efficiency of BF-GAN in generating bubbly flow images.

Table 2. Comparison of AI indicators between BF-GAN and conventional GAN.

| All conditions | FID ↓ | KID ↓ | P ↑ | R ↑ | PPL ↓ | EQ-T int ↑ | EQ-T frac ↑ | EQ-R ↑ | IS ↑ |
|---|---|---|---|---|---|---|---|---|---|
| BF-GAN | **13.261** | **0.003** | **0.674** | **1.097E-03** | **1.896** | **48.766** | **43.980** | **17.910** | **2.749** |
| Conventional GAN | 32.610 | 0.013 | 0.414 | 1.799E-05 | 2.881 | 44.838 | 43.099 | 17.517 | 2.467 |

**Fig. 15** visualizes the workflow of the generator in BF-GAN. The process starts with the input combining Gaussian noise and a feature matrix, labeled as $j_g$ and $j_f$, to initiate the generation process. The input is successively processed through 14-layer CNN, each with specific kernel sizes and channel numbers as indicated below each image. The architecture gradually refines the initial noisy input into structured images that resemble patterns and complex textural details of bubbly flow.

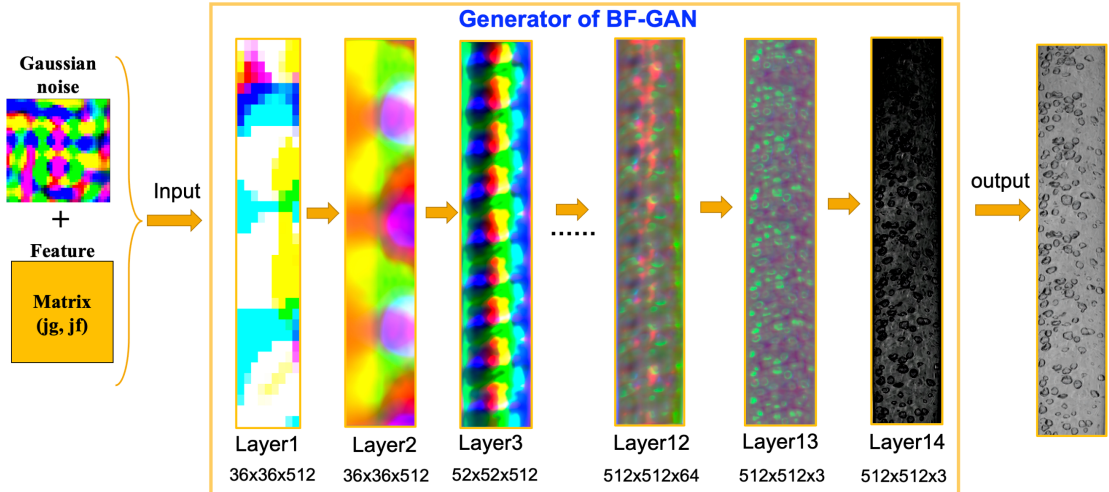

Fig. 15. Visualization of the generator of BF-GAN.

In the supplementary video, a demonstration video was produced to visualize how does the BF-GAN generates the images during the smooth transition between $j_g$ and $j_f$. Initially, images with various conditions were generated by inputting different and continuous values for $j_g$ and $j_f$, along with a random seed. Subsequently, interpolation of these generated bubbly flow images was performed to create the video.

### 3.3 Quantitative assessment of BF-GAN from the perspective of image correspondence

In this section, the image correspondence was evaluated between 525,000 bubbly flow images generated by BF-GAN and 278,000 experimental images. For each comparison, 5000 images were generated by BF-GAN, resulting in a total of 105 comparisons. Five image correspondence indicators—luminance, contrast, magnitude, homogeneity, and correlation—were utilized to comprehensively compare the images from real experiments with those generated by BF-GAN.

### 3.3.1 Luminance
Luminance refers to the average light intensity of an image, which can be expressed as the mean value of pixel intensities. In the evaluation of bubbly flow images, luminance aids in understanding the visibility of the bubbles and the background lighting. The formula for calculating luminance is as follows:

$$\text{Luminance} = \frac{1}{n}\sum_{i=1}^{n} \text{img}_i \qquad (7)$$

where $n$ is the total number of pixels in the image, and $\text{img}_i$ represents the intensity value of the $ith$ pixel. **Table 3** presents the comparative luminance results of these bubbly flow images. The mean of the absolute mean relative error (MAMRE) of luminance in 105 experiments is 2.22%. It indicates that the luminance of the generated images closely approximates that of the experimental images. The MRE map of luminance is shown in **Fig. 16.**

**Table 3. Comparative luminance results.**

| Num | $j_g$ [m/s] | $j_f$ [m/s] | Luminance of experiment images | Luminance of BF-GAN | MRE |
|---|---|---|---|---|---|
| 1 | 0.029 | 0.215 | 100.67 | 99.49 | -1.17% |
| 2 | 0.076 | 0.217 | 96.83 | 91.68 | -5.32% |
| 3 | 0.024 | 0.322 | 100.20 | 100.05 | -0.15% |
| … | … | … | … | … | … |
| 103 | 0.095 | 0.375 | 97.00 | 95.94 | -1.09% |
| 104 | 0.152 | 0.384 | 91.30 | 89.30 | -2.20% |
| 105 | 0.233 | 0.395 | 85.88 | 83.48 | -2.79% |

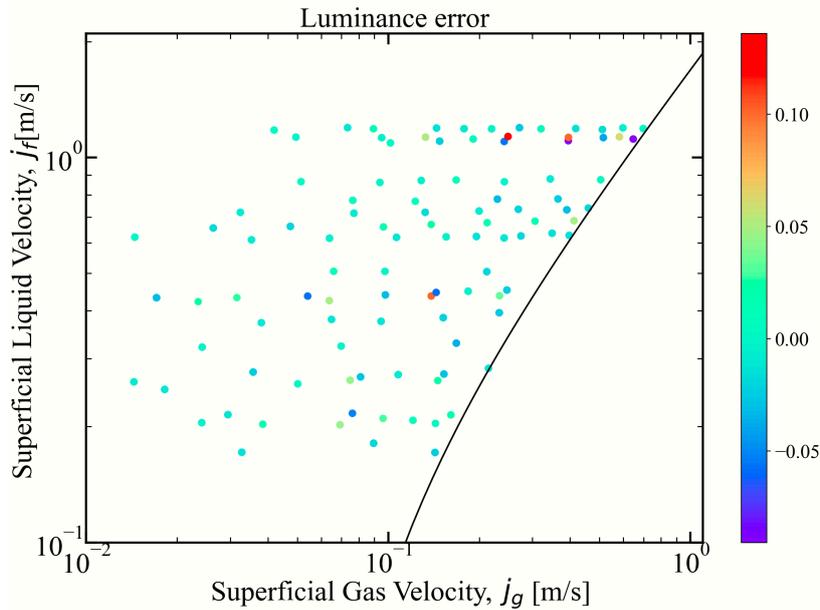

Fig. 16. MRE map of luminance between BF-GAN and experimental images.

### 3.3.2 Contrast

Contrast refers to the degree of difference between its brightest and darkest areas, and is typically estimated by calculating the standard deviation of pixel intensities within the image. High contrast indicates a clearer outline between the bubbles and background. The formula for calculating contrast is as follows:

$$\text{Contrast} = \sqrt{\frac{1}{n}\sum_{i=1}^{n}(\text{img}_i - \mu)^2} \qquad (8)$$

where, $\mu$ represent the mean value of all pixel intensities in the image. **Table 4** presents the comparative results for contrast. The MAMRE of contrast in 105 experiments is 2.93%. Luminance and contrast are directly related to the way light is reflected and refracted on bubble surfaces. The MRE map of contrast is illustrated in **Fig. 17**.

**Table 4. Comparative contrast results.**

| Num | $j_g$ [m/s] | $j_f$ [m/s] | Contrast of experiment images | Contrast of BF-GAN | MRE |
|---|---|---|---|---|---|
| 1 | 0.029 | 0.215 | 24.53 | 22.96 | -6.40% |
| 2 | 0.076 | 0.217 | 29.11 | 27.40 | -5.86% |
| 3 | 0.024 | 0.322 | 22.71 | 21.32 | -6.11% |
| … | … | … | … | … | … |
| 103 | 0.095 | 0.375 | 27.34 | 26.74 | -2.19% |
| 104 | 0.152 | 0.384 | 28.76 | 28.56 | -0.69% |
| 105 | 0.233 | 0.395 | 29.22 | 28.87 | -1.18% |

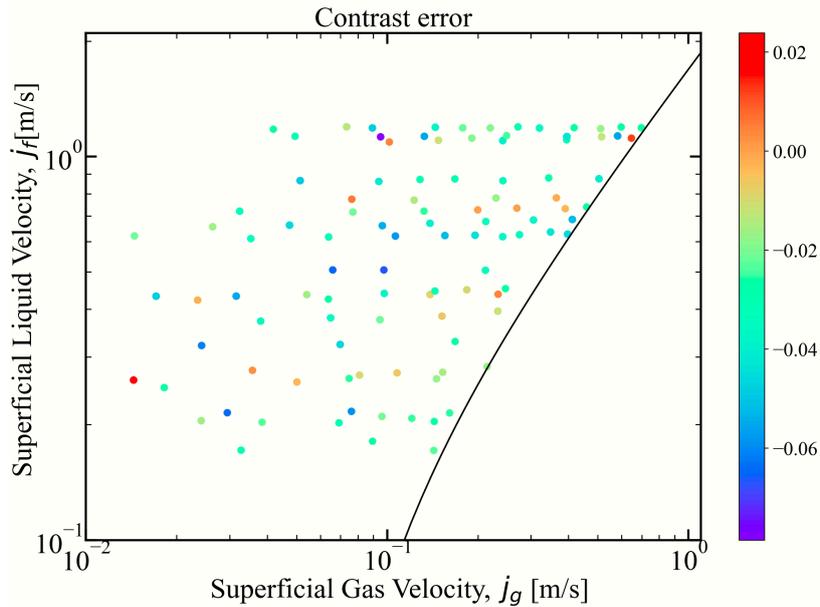

Fig. 17. MRE map of contrast between BF-GAN and experimental images.

### 3.3.3 Magnitude

The magnitude of the gradient is commonly used to measure the intensity of edges within an image [39]. In the present study, edge detection utilizing the Sobel operator was employed to assess the magnitude of images. Comparing magnitude assists in determining whether the edges of bubbles in images generated by BF-GAN resemble those in experimental images. The calculation of magnitude includes the bubbles breakup and coalescence. The formula for calculating magnitude is as follows:

$$\text{Magnitude} = \sqrt{\text{sobel}(x)^2 + \text{sobel}(y)^2} \tag{9}$$

here, $\text{sobel}(\cdot)$ denotes the result of applying the Sobel operator to the image in the $x$ or $y$ direction. **Table 5** presents the comparative results for the magnitude of these bubbly flow images. The MAMRE of magnitude in 105 experiments is 24.74%. The MRE map of magnitude is shown in **Fig. 18.**

Table 5. Comparative magnitude results.

| Num | $j_g$ [m/s] | $j_f$ [m/s] | Magnitude of experiment images | Magnitude of BF-GAN | MRE |
|---|---|---|---|---|---|
| 1 | 0.029 | 0.215 | 623.32 | 829.17 | 33.03% |
| 2 | 0.076 | 0.217 | 827.91 | 1021.86 | 23.43% |
| 3 | 0.024 | 0.322 | 575.21 | 744.48 | 29.43% |
| … | … | … | … | … | … |
| 103 | 0.095 | 0.375 | 808.81 | 1024.26 | 26.64% |
| 104 | 0.152 | 0.384 | 848.99 | 1072.37 | 26.31% |
| 105 | 0.233 | 0.395 | 875.79 | 1055.57 | 20.53% |

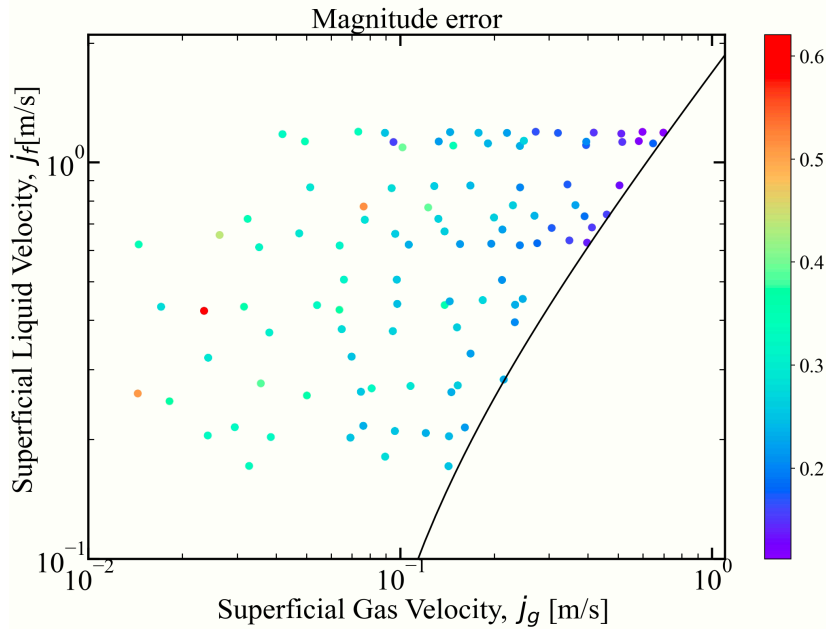

**Fig. 18. MRE map of magnitude between BF-GAN and experimental images.**

### 3.3.4 Homogeneity

Homogeneity describes the consistency or smoothness within local regions of an image [40]. Within the context of the Gray Level Co-occurrence Matrix (GLCM), it measures the intensity near the diagonal of the matrix, reflecting the proximity of similar gray levels in the image. In bubbly flow images, high homogeneity indicates a uniform distribution of bubbles without excessive noise. This indicator is utilized to evaluate the visual smoothness and realism of bubbly flow images generated by BF-GAN. The formula for calculating homogeneity is presented as follows:

$$\text{Homogeneity} = \sum_{i,j} \frac{1}{1+(i-j)^2} \text{glcm}_{i,j} \qquad (10)$$

here, $i$ and $j$ are the pixel intensity values, $\text{glcm}_{i,j}$ is the frequency of occurrence of pixel with intensity $i$ adjacent to a pixel with intensity $j$. **Table 6** lists the comparative results for homogeneity. The MAMRE in all 105 experiments is 21.34%. **Fig. 19** illustrates the MRE homogeneity map.

**Table 6. Comparative homogeneity results.**

| Num | $j_g$ [m/s] | $j_f$ [m/s] | Homogeneity of experiment images | Homogeneity of BF-GAN | MRE |
|---|---|---|---|---|---|
| 1 | 0.029 | 0.215 | 0.35 | 0.28 | -18.70% |
| 2 | 0.076 | 0.217 | 0.27 | 0.21 | -19.85% |
| 3 | 0.024 | 0.322 | 0.36 | 0.31 | -15.98% |
| … | … | … | … | … | … |
| 103 | 0.095 | 0.375 | 0.27 | 0.21 | -21.04% |
| 104 | 0.152 | 0.384 | 0.24 | 0.19 | -23.48% |
| 105 | 0.233 | 0.395 | 0.22 | 0.17 | -21.46% |

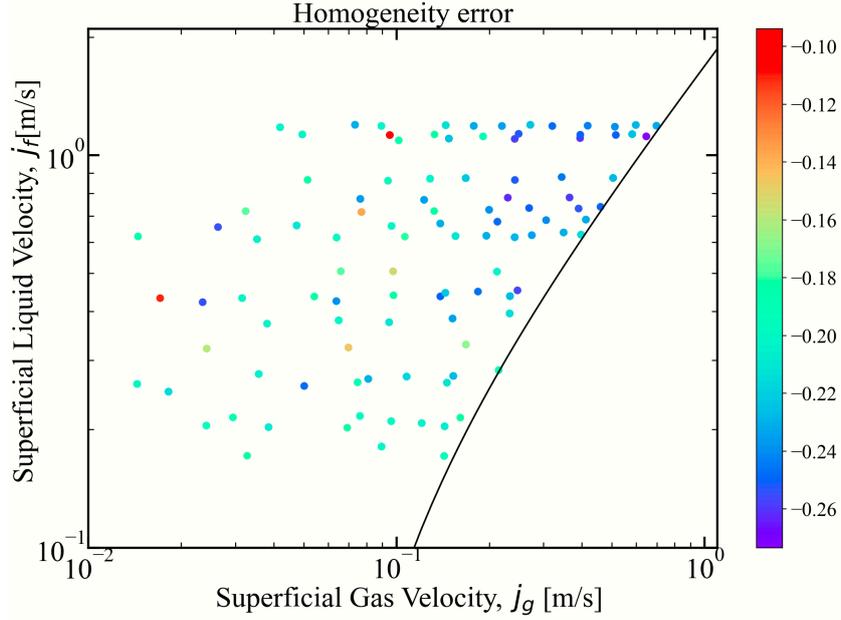

**Fig. 19. MRE map of homogeneity between BF-GAN and experimental images.**

**3.3.5 Correlation**

Correlation quantifies the linear relationships between different gray levels in the image [41]. This indicator reflects the structure and texture patterns of bubbly flow, including the spatial distribution and position of bubbles under transitional flow regimes. The formula for correlation is given as:

$$\text{Correlation} = \sum_{i,j} \frac{(i-\mu_i)(j-\mu_j)\text{glcm}_{i,j}}{\sigma_i \sigma_j} \tag{11}$$

where, $\mu_i$ and $\mu_j$ are the mean values of the gray levels $i$ and $j$, respectively. $\sigma_i$ and $\sigma_j$ are the standard deviations of gray levels $i$ and $j$, respectively.

**Table 7** presents the comparative results for correlation. The MAMRE in all 105 experiments is 13.59%. **Fig. 20** depicts the MRE map of correlation.

**Table 7. Comparative correlation results.**

| Num | $j_g$ [m/s] | $j_f$ [m/s] | Correlation of experiment images | Correlation of BF-GAN | MRE |
|---|---|---|---|---|---|
| 1 | 0.029 | 0.215 | 0.910 | 0.789 | -13.30% |
| 2 | 0.076 | 0.217 | 0.906 | 0.805 | -11.19% |
| 3 | 0.024 | 0.322 | 0.903 | 0.788 | -12.79% |
| … | … | … | … | … | … |
| 103 | 0.095 | 0.375 | 0.904 | 0.788 | -12.76% |
| 104 | 0.152 | 0.384 | 0.908 | 0.801 | -11.81% |
| 105 | 0.233 | 0.395 | 0.908 | 0.809 | -10.92% |

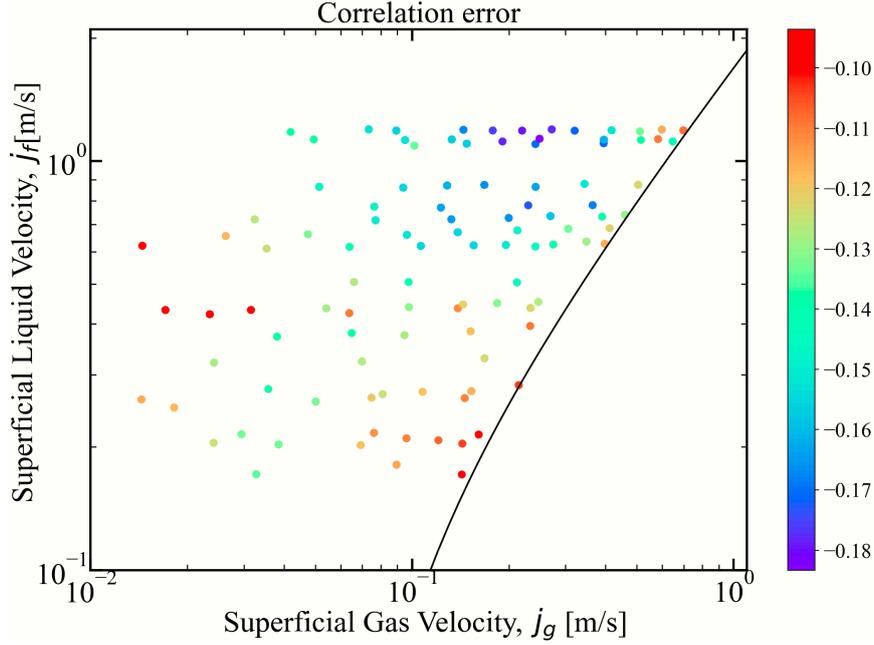

**Fig. 20. MRE map of correlation between BF-GAN and experimental images.**

**Table 8** lists the averages, average MREs, and maximum MREs for five indicators: luminance, contrast, magnitude, homogeneity, and correlation.

Table 8. Comparative results for image correspondence of the BF-GAN.

| Indicators | Luminance | Contrast | Magnitude | Homogeneity | Correlation |
|---|---|---|---|---|---|
| Mean BF-GAN | 91.106 | 25.192 | 969.211 | 0.219 | 0.782 |
| Mean EXP. | 91.517 | 25.925 | 777.006 | 0.277 | 0.904 |
| MAMRE | 2.22% | 2.93% | 24.74% | 21.34% | 13.59% |
| Max. absolute MRE | 13.59% | 7.86% | 62.07% | 27.34% | 18.33% |

In general, while the generated images of BF-GAN demonstrate high consistency with the experimental bubbly flow images in terms of image correspondence across all conditions, the MRE maps clearly indicate that the highest errors in magnitude and homogeneity are predominantly concentrated in regions with high $j_g$ and $j_f$ areas. This may be attributed to the extensive overlap of bubbles under these conditions, which complicates the outline of bubble edges. Additionally, the scattering of light by numerous bubbles alters the perceived uniformity. These factors collectively reduce BF-GAN's ability to accurately generate these features.

### 3.4 Quantitative assessment of BF-GAN from the perspective of two-phase flow parameters

In this section, the two-phase flow parameters of the bubbly flow images generated by BF-GAN were extracted using a bubble detection model and then compared with experimental images to validate the accuracy of the generated images' two-phase flow parameters. Considering that the bubble detection model was developed specifically for areas with low void fractions, only the 38 red points shown in **Fig. 21** were utilized for the comparison.

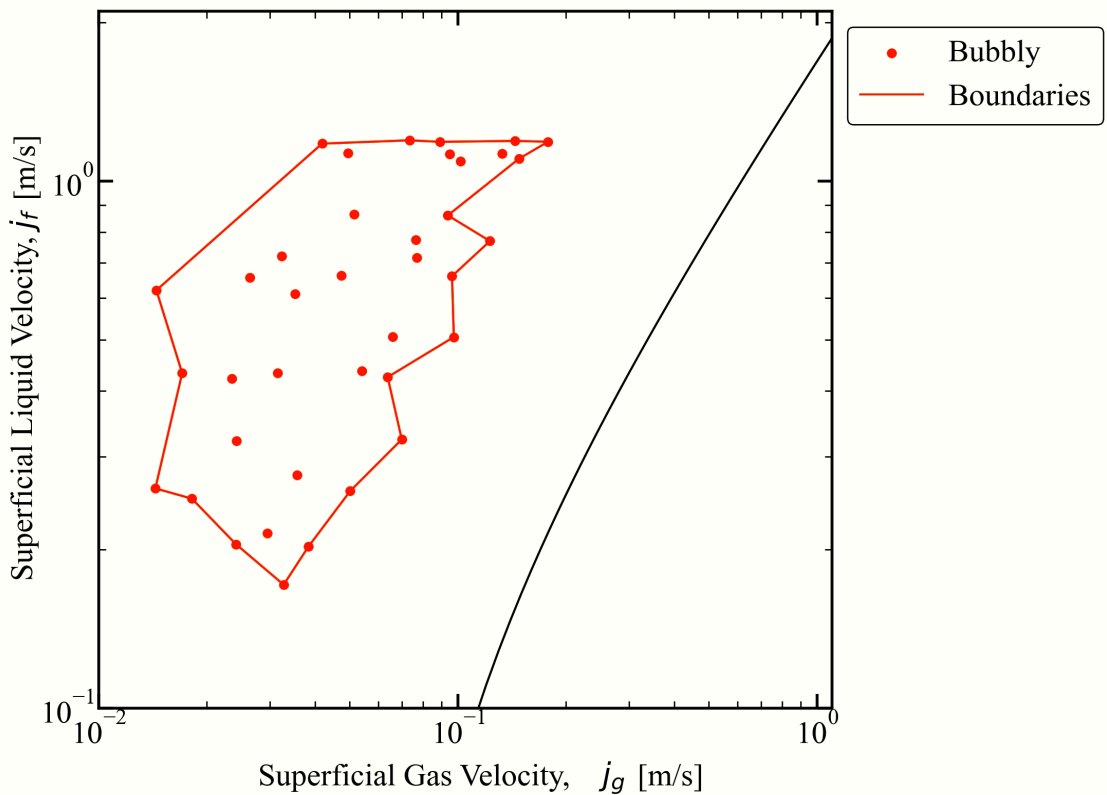

**Fig. 21. Validation area of two-phase flow parameter for the BF-GAN.**

Bubbly flow images generated by BF-GAN were detected using the bubble detection model described in Section 2.3. The detection results are illustrated in **Fig. 22 (a)**, where the bounding area of each individual bubble was extracted, as shown in **Fig. 22 (b)**. In the present study, each bubble was assumed to be a three-dimensional ellipsoid with axes $a, b$, and $c$. The $c$ is calculated using the following formula:

$$c = \frac{a+b}{2} \qquad (12)$$

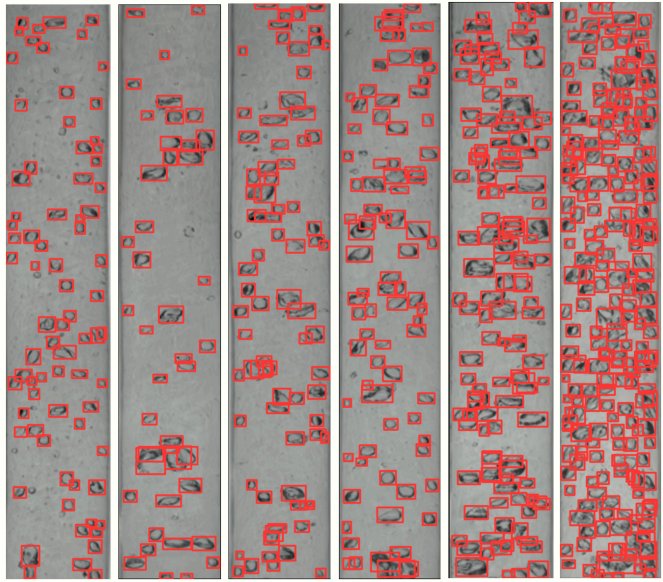

(a). Sample results of the bubble detection model.

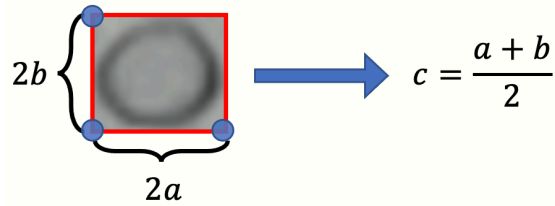

(b). Extraction of the semi-axes of the ellipsoid bubble.

Fig. 22. Application of the bubble detection model with the BF-GAN.

### 3.4.1 Void fraction

The void fraction represents the volumetric fraction of the gas phase within a flow channel. It is dimensionless and varies between 0 and 1, directly influencing the flow regime and heat transfer characteristics. Mathematically, the void fraction $\alpha$ can be expressed:

$$\alpha = \frac{V_{gas}}{V_{total}} \qquad (13)$$

where, $V_{gas}$ represents the gas phase volume, and $V_{total}$ is the total pipe volume.

The extraction of void fraction by BF-GAN is defined as:

$$\alpha = \frac{V_{bubble}}{V_{total}} = \frac{\Sigma V_n}{\pi D^2 L/4} \qquad (14)$$

$$V = \frac{3}{4}\pi abc \qquad (15)$$

here, $V_{total}$ represents the total volume of the pipeline, where $D$ is the diameter of the pipe and $L$ is the length of the pipe. $V$ denotes the volume of each individual bubble.

The actual void fraction values were obtained based on the experimental images. **Table 9** presents the comparative results of the void fraction between the BF-GAN and the experimental images, with a MAMRE for the 38 comparisons being 14.64%. **Fig. 23** illustrates the MRE map of the void fraction between BF-GAN and experimental images.

**Table 9. Comparative results for void fraction [-].**

| Num | $j_g$ [m/s] | $j_f$ [m/s] | Void faction of experimental images | Void faction of BF-GAN | MRE |
|---|---|---|---|---|---|
| 1 | 0.029 | 0.215 | 0.134 | 0.088 | -34.04% |
| 2 | 0.024 | 0.322 | 0.080 | 0.086 | 6.40% |
| 3 | 0.070 | 0.324 | 0.292 | 0.256 | -12.30% |
| … | … | … | … | … | … |
| 36 | 0.094 | 0.861 | 0.066 | 0.063 | -5.65% |
| 37 | 0.014 | 0.621 | 0.013 | 0.013 | -3.94% |
| 38 | 0.035 | 0.611 | 0.044 | 0.034 | -22.71% |

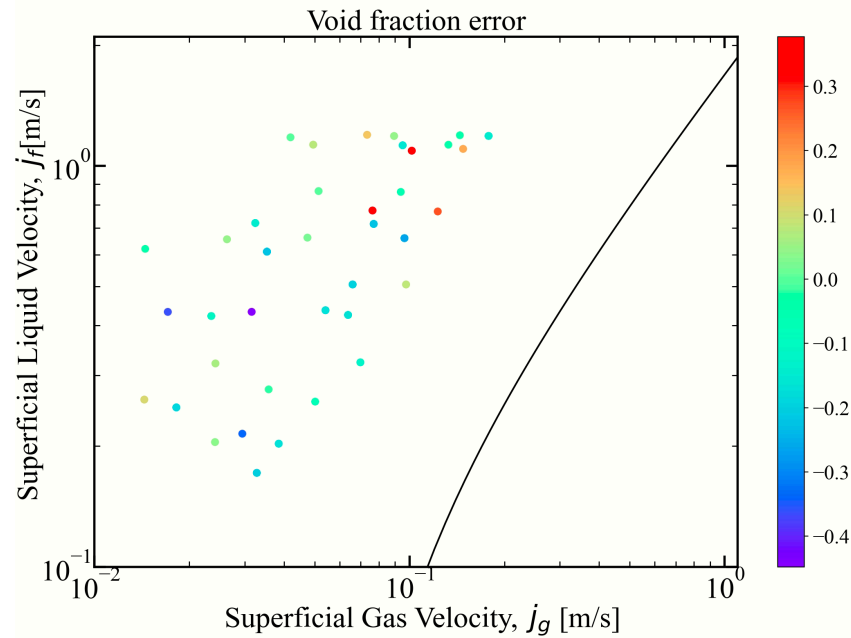

**Fig. 23. MRE map of void faction between BF-GAN and experimental images.**

### 3.4.2 Aspect ratio
Aspect ratio refers to the ratio of characteristic lengths perpendicular to the flow direction to those along the flow direction. It is dimensionless and affects the phase distribution and pressure drop, thereby impacting the thermal performance and operational efficiency of heat exchangers and micro-reactors. Its mathematical definition can be given as follows:

$$E = \frac{MinorAxisLength}{MajorAxisLength} = \frac{2a}{2b} \tag{16}$$

Through the bubble detection model, $a$ and $b$ were readily obtained, as shown in **Fig. 22(b)**. **Table 10** displays the comparative results for the aspect ratios, with a MAMRE of 2.31%. The MRE map for the aspect ratio between BF-GAN and experimental images is depicted in **Fig. 24**.

**Table 10. Comparative results for aspect ratio [-].**

| Num | $j_g$ [m/s] | $j_f$ [m/s] | Aspect ratio of experimental images | Aspect ratio of BF-GAN | MRE |
|---|---|---|---|---|---|
| 1 | 0.029 | 0.215 | 0.705 | 0.702 | -0.41% |
| 2 | 0.024 | 0.322 | 0.703 | 0.683 | -2.86% |
| 3 | 0.070 | 0.324 | 0.730 | 0.708 | -2.99% |
| … | … | … | … | … | … |
| 36 | 0.094 | 0.861 | 0.762 | 0.744 | -2.33% |
| 37 | 0.014 | 0.621 | 0.741 | 0.745 | 0.56% |
| 38 | 0.035 | 0.611 | 0.728 | 0.718 | -1.46% |

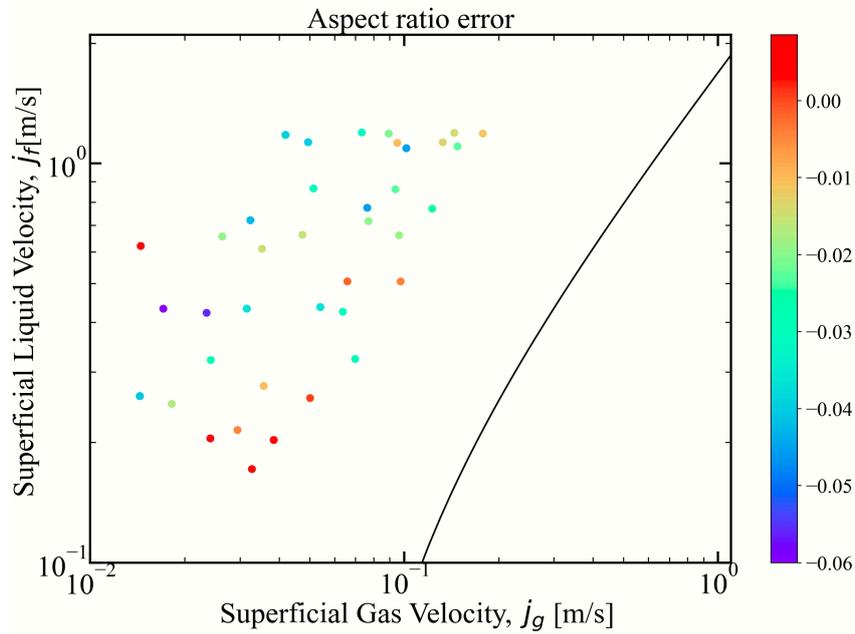

**Fig. 24. MRE map of aspect ratio between BF-GAN and experimental images.**

### 3.4.3 Sauter mean diameter
Sauter mean diameter (SMD) represents the diameter of a sphere that has the same volume/surface area ratio as the average of the droplets or bubbles in the dispersion. SMD is used in the optimization of spray processes, as it influences the surface area available for mass and heat transfer processes between phases. Mathematically, the SMD ($D_{SM}$) is calculated:

$$D_{SM} = \frac{6}{n} \frac{\sum_{i=1}^{n} V_n}{\sum_{i=1}^{n} A_n} \tag{17}$$

$$A \approx 4\pi \sqrt[p]{\frac{a^p b^p + a^p c^p + b^p c^p}{3}} \tag{18}$$

where, $A$ represents the estimated surface area of an ellipsoid bubble, calculated using Knud Thomsen's approximation. The surface area value derived from this approximation exhibits the lowest relative error when $p=1.6075$ [42, 43]. **Table 11** presents the comparative results for SMD, with a MAMRE of 5.36%. The MRE map for SMD is depicted in **Fig. 25**.

**Table 11. Comparative results for Sauter mean diameter [m].**

| Num | $j_g$ [m/s] | $j_f$ [m/s] | Sauter mean diameter of experimental images | Sauter mean diameter of BF-GAN | MRE |
|---|---|---|---|---|---|
| 1 | 0.029 | 0.215 | 3.920E-03 | 4.041E-03 | 3.09% |
| 2 | 0.024 | 0.322 | 3.652E-03 | 3.953E-03 | 8.24% |
| 3 | 0.070 | 0.324 | 4.351E-03 | 4.474E-03 | 2.81% |
| … | … | … | … | … | … |
| 36 | 0.094 | 0.861 | 3.618E-03 | 3.786E-03 | 4.64% |
| 37 | 0.014 | 0.621 | 3.749E-03 | 3.934E-03 | 4.92% |
| 38 | 0.035 | 0.611 | 3.783E-03 | 3.796E-03 | 0.34% |

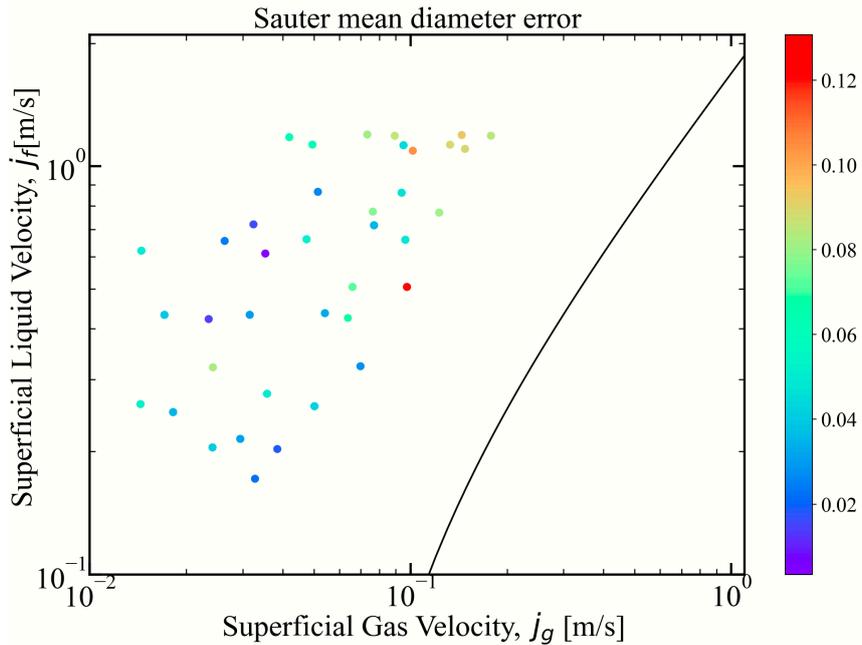

**Fig. 25. MRE map of Sauter mean diameter between BF-GAN and experimental**

### 3.4.4 IAC

Interfacial area concentration (IAC) quantifies the total interfacial area per unit volume between two phases. High IAC values typically indicate enhanced transfer rates between the two phases. Its mathematical definition can be given as follows:

$$\alpha_i = \frac{\sum_{i=1}^{n} A_n}{V_{total}} \tag{19}$$

**Table 12** provides the comparative analysis of the IAC, with a MAMRE of 8.03%. The MRE map for the IAC is depicted in **Fig. 26**.

**Table 12. Comparative results for IAC [1/m].**

| Num | $j_g$ [m/s] | $j_f$ [m/s] | IAC of experimental images | IAC of BF-GAN | MRE |
|---|---|---|---|---|---|
| 1 | 0.029 | 0.215 | 69.458 | 66.302 | -4.54% |
| 2 | 0.024 | 0.322 | 64.164 | 60.246 | -6.11% |
| 3 | 0.070 | 0.324 | 119.575 | 115.977 | -3.01% |
| … | … | … | … | … | … |
| 36 | 0.094 | 0.861 | 82.730 | 84.626 | 2.29% |
| 37 | 0.014 | 0.621 | 15.780 | 14.925 | -5.42% |
| 38 | 0.035 | 0.611 | 45.843 | 43.149 | -5.88% |

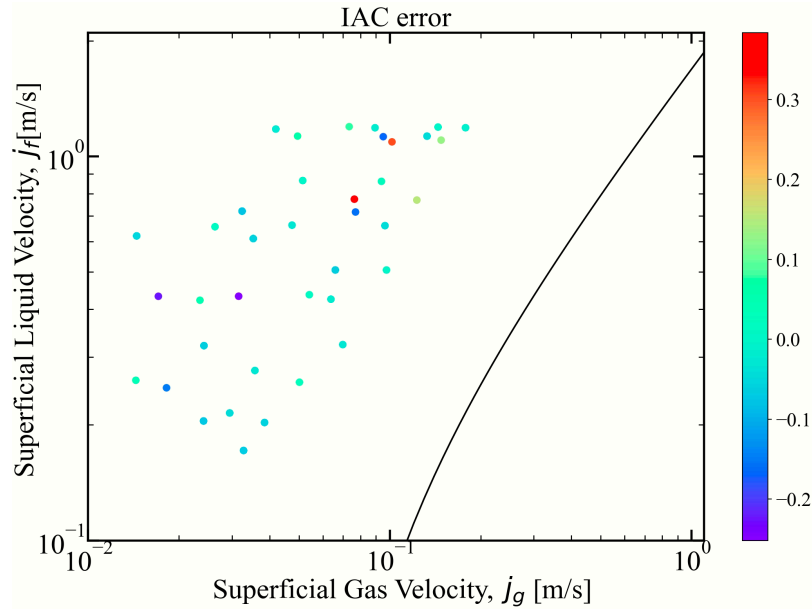

**Fig. 26. MRE map of Interfacial area concentration between BF-GAN and experimental images.**

The average and maximum MRE for the two-phase flow parameters obtained using the BF-GAN are presented in **Table 13**. Overall, the bubbly flow images generated by BF-GAN demonstrates that two-phase flow parameters are comparable to experimental images. However, as evident from the MRE maps, areas with significant errors are predominantly localized around the boundaries of these 38 points, particularly in low and high $j_g$ and $j_f$ area. This may lead to detection errors due to the presence of cap bubbles, which do not conform to the standard ellipsoidal shape. In high $j_g$ and $j_f$ area, the extensive overlap and coverage of bubbles may result in the loss or inaccuracy of some bounding box information. It is important to emphasize that, although the maximum error for the void faction is 44.76%, the experimental image value at this point is 0.026, while the BF-GAN value is 0.046. The relatively large error is attributed to the small denominator.

Table 13. Comparative results for two-phase flow parameters of the BF-GAN.

| BF-GAN | Void faction | Aspect ratio | SMD | IAC |
|---|---|---|---|---|
| MAMRE | 14.64% | 2.28% | 5.23% | 7.81% |
| Max. absolute MRE | 44.76% | 6.00% | 13.07% | 38.39% |

It is worth noting that the bubble detection model utilized in the present study was also developed based on the BF-GAN's dataset. Therefore, the combined application of BF-GAN with the bubble detection model enhances the efficiency significantly. The bubble detection model not only facilitates the easy extraction of two-phase flow parameters for each bubble but also for the entire bubbly flow image, as illustrated in **Fig. 27**. The models and data developed in the present study have been made openly available on GitHub. For further details, please refer to our GitHub repository.

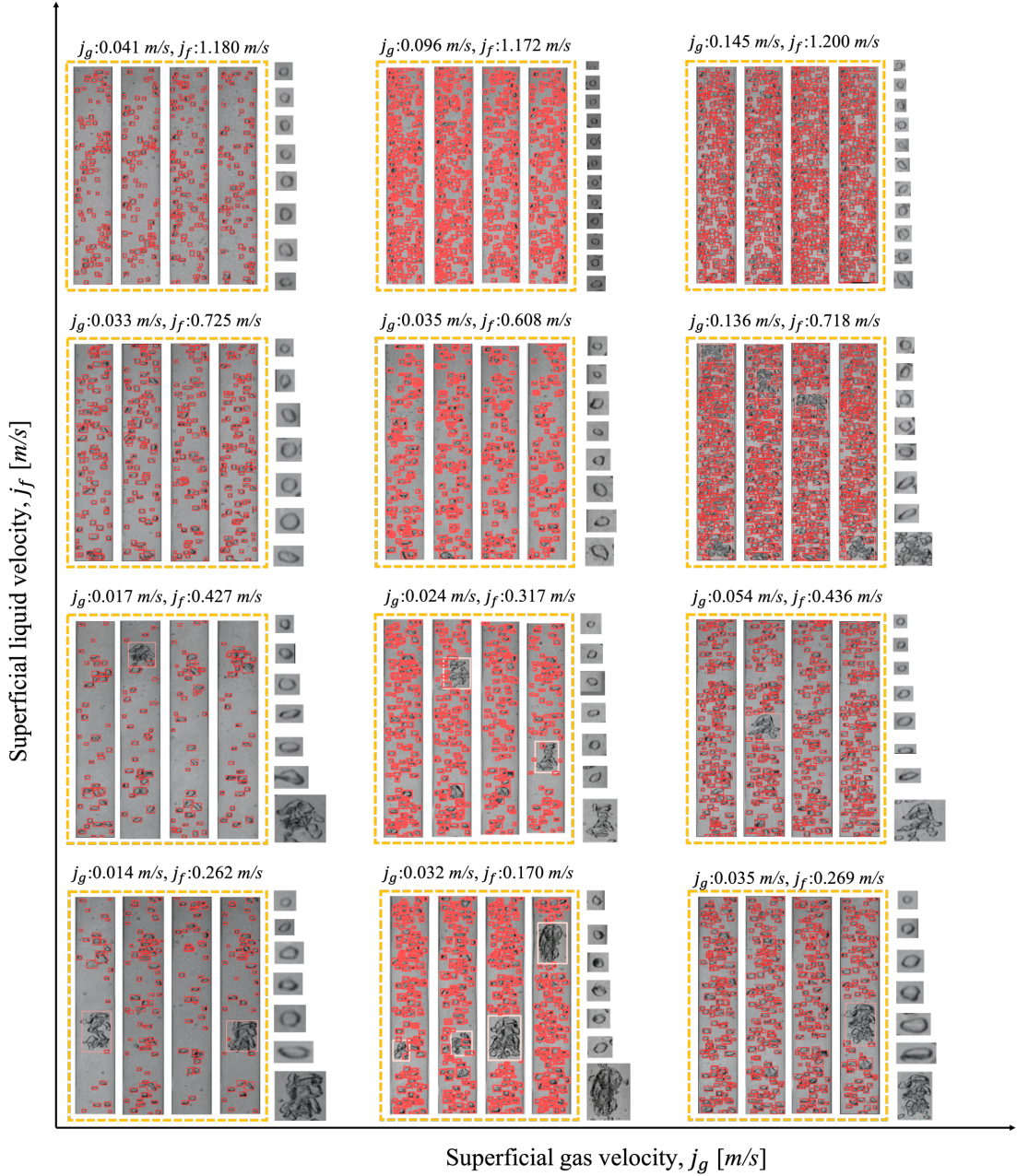

**Fig. 27. Detection results of BF-GAN generative images, and parameter extraction of individual bubbles.**

# 4 Conclusion and ongoing work

A generative AI architecture, BF-GAN, has been developed for generating high-quality bubbly flow images. A dataset comprising 278,000 images was collected from 105 different experiments with varying $j_g$ and $j_f$, serving as the training set. An NVIDIA generator and a multi-scale loss function were developed to enhance the performance of BF-GAN in generating bubbly flow images. The efficacy of BF-GAN was validated across generative AI indicators, image correspondence, and two-phase flow parameters, confirming its capability to generate high-fidelity bubbly flow images under various $j_g$ and $j_f$ flow conditions. By integrating a YOLO-based bubble detection model with conditional BF-GAN, automated generation and parameter extraction of bubbly flow images have been achieved.

Ongoing work is divided into two main areas. First, to more comprehensively and clearly generate gas-liquid two-phase flow images, three flow patterns—slug, churn, and annular—will also be included within the scope of the generative AI. A diffusion model will be employed for training generative AI across all $j_g$ and $j_f$ conditions, extending to 1024 pixels resolution. Second, inspired by the latest advancements in text-to-video and image-to-video AI generative technologies, the generation of two-phase flow videos from textual or image inputs will be explored. These videos will not only facilitate the extraction of static parameters but will also allow for the extraction of time-dependent information such as bubble velocity and interface changes. Related works will be reported in the near future.

**Supplementary Video 1**
In the supplementary video, a demonstration video was produced to visualize how does the BF-GAN generates the images during the smooth transition between $j_g$ and $j_f$. Initially, images with various conditions were generated by inputting different and continuous values for $j_g$ and $j_f$, along with a random seed. Subsequently, interpolation of these generated bubbly flow images was performed to create the video.
https://www.dropbox.com/scl/fi/dh4ikfay6bxyh5jlkhero/bubbly.mp4?rlkey=62yuqt2i37v7c3z3ly9f7bzir&st=8ou89r0j&dl=0

**Supplementary File 1**
The File (BF-GAN_results.xlsx) includes the comparison of image correspondence and two-phase flow parameters under all conditions, as well as the benchmark conditions.

**Appendix 1: Release of bubbly flow benchmark datasets**
To help reduce usage costs, red boundaries were delineated based on the current distribution of the dataset, as shown in **Fig. 28**. Subsequently, increments of 5% in $j_g$ and $j_f$ were applied, denoted by blue points, resulting in a total of 2080 $j_g$ and $j_f$ conditions. Each condition was generated by six BF-GAN models, generating a total of 3000 images per condition (500 images per model). Consequently, a dataset comprising 6.24 million bubbly flow images was constructed, corresponding to the blue-point $j_g$ and $j_f$ conditions.

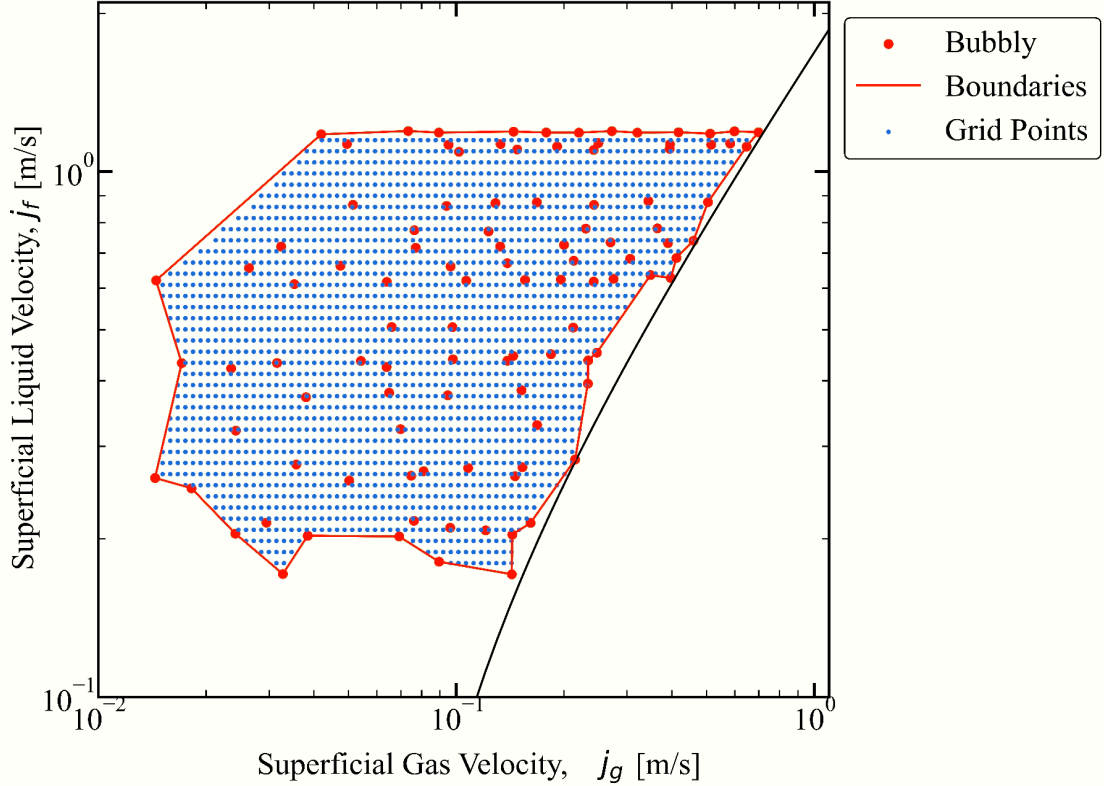

**Fig. 28. Conditions of bubbly flow benchmark datasets**

**Appendix 2: Comparison of the BF-GAN and experimental measurement, empirical correlations.**

To further validate the authenticity of bubbly flow images generated by BF-GAN, the void fraction of the BF-GAN-generated bubbly flow images was compared with a wire mesh sensor measurement. Additionally, the aspect ratio, Sauter mean diameter, and interfacial area concentration were compared with empirical correlations.

**1. Void fraction**

**Table 14** presents the comparative results of the void fraction between the BF-GAN and a wire mesh sensor, with a MAMRE for the 39 comparisons being 30.62%. **Fig. 29** illustrates the MRE map of the void fraction generated by the BF-GAN.

**Table 14. Comparative results for void fraction [-].**

| Num | $j_g$ [m/s] | $j_f$ [m/s] | Void faction of wire mesh sensor | Void faction of BF-GAN | MRE |
|---|---|---|---|---|---|
| 1 | 0.029 | 0.215 | 0.122 | 0.088 | -27.76% |
| 2 | 0.024 | 0.322 | 0.177 | 0.086 | -51.58% |
| 3 | 0.070 | 0.324 | 0.260 | 0.256 | -1.43% |
| … | … | … | … | … | … |
| 36 | 0.094 | 0.861 | 0.042 | 0.063 | 47.48% |
| 37 | 0.014 | 0.621 | 0.011 | 0.013 | 20.51% |

| 38 | 0.035 | 0.611 | 0.025 | 0.034 | 34.99% |

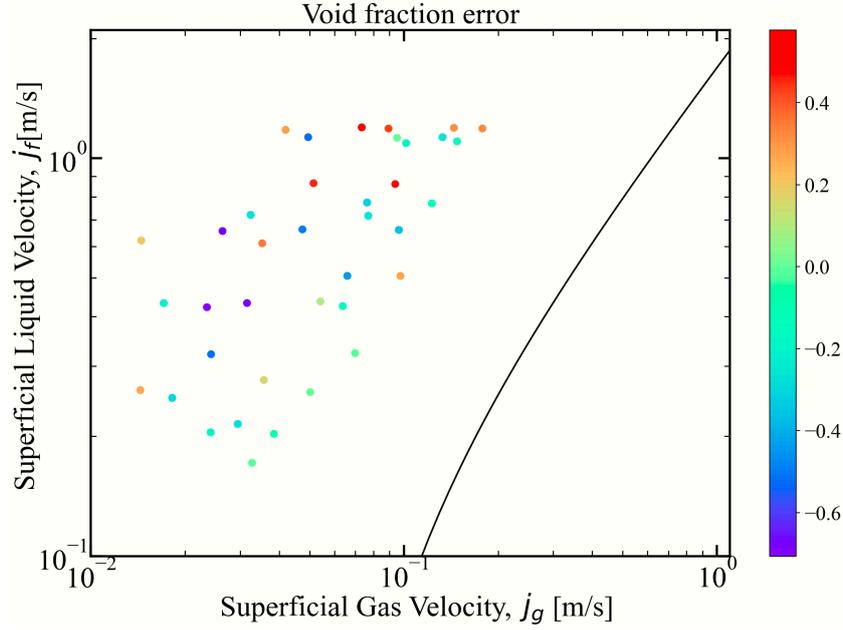

**Fig. 29. MRE map of void faction between BF-GAN and wire mesh sensor.**

**2. Aspect ratio**

In the present study, Besagni's correlation [44] was utilized to evaluate the aspect ratio of bubbly flow images generated by BF-GAN, as presented in Eq. (20).

$$E = \frac{1}{[1+0.45(EoRe)]^{0.08}} \quad (20)$$

$$Eo = \frac{g(\rho_L - \rho_G)d_{eq}^2}{\sigma} \quad (21)$$

$$Re = \frac{j_f \rho_l D}{\mu_l} \quad (22)$$

where, $Eo$ is Eötvös number, $Re$ is Reynolds number. $\rho_l$, $\rho_g$, $g$, $d_{eq}$, $\sigma$, and $\mu_l$, denote the density of the liquid, the density of the gas, the gravitational acceleration, the equivalent diameter of the bubble, the surface tension, and the dynamic viscosity of the liquid, respectively.

**Table 15** displays the comparative results for the aspect ratios, with a MAMRE of 18.39%. The MRE map for the aspect ratio is depicted in **Fig. 30**.

**Table 15. Comparative results for aspect ratio [-].**

| Num | $j_g$ [m/s] | $j_f$ [m/s] | Aspect ratio of empirical correlation | Aspect ratio of BF-GAN | MRE |
|---|---|---|---|---|---|
| 1 | 0.029 | 0.215 | 0.649 | 0.702 | 8.20% |
| 2 | 0.024 | 0.322 | 0.662 | 0.683 | 3.12% |
| 3 | 0.070 | 0.324 | 0.596 | 0.708 | 18.95% |
| … | … | … | … | … | … |
| 36 | 0.094 | 0.861 | 0.598 | 0.744 | 24.47% |
| 37 | 0.014 | 0.621 | 0.689 | 0.745 | 8.08% |
| 38 | 0.035 | 0.611 | 0.646 | 0.718 | 11.08% |

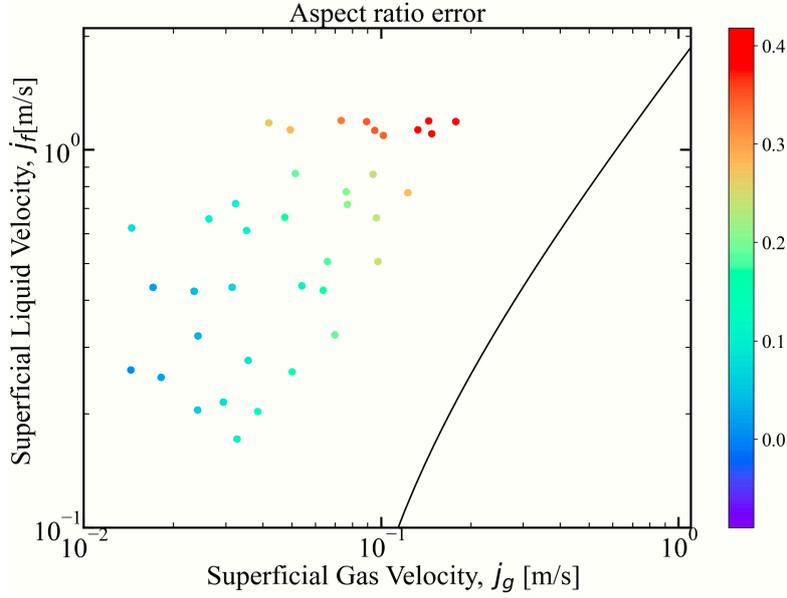

**Fig. 30.** MRE map of aspect ratio between BF-GAN and empirical correlation.

### 3. Sauter mean diameter

Hibiki's empirical correlation [45] for SMD was employed to evaluate the performance of BF-GAN. **Table 16** presents the comparative results for SMD, with a MAMRE of 21.81%. The MRE map for SMD is depicted in **Fig. 31**.

$$D_{SMD} = 1.63 \left(\frac{Lo}{D}\right)^{-0.335} \alpha^{0.170} N_{Re_b}^{-0.239} \left(\frac{\rho_l}{\rho_g}\right)^{0.138} Lo \quad (23)$$

here, $Lo$, $\alpha$, and $N_{Re_b}$ are the Laplace length, the void fraction, and the bubble Reynolds number, respectively.

**Table 16. Comparative results for Sauter mean diameter [m].**

| Num | $j_g$ [m/s] | $j_f$ [m/s] | Sauter mean diameter of empirical correlation | Sauter mean diameter of BF-GAN | MRE |
|---|---|---|---|---|---|
| 1 | 0.029 | 0.215 | 5.698E-03 | 4.041E-03 | -29.08% |
| 2 | 0.024 | 0.322 | 5.555E-03 | 3.953E-03 | -28.84% |
| 3 | 0.070 | 0.324 | 5.121E-03 | 4.474E-03 | -12.63% |
| … | … | … | … | … | … |

| 36 | 0.094 | 0.861 | 4.446E-03 | 3.786E-03 | -14.84% |
| 37 | 0.014 | 0.621 | 5.333E-03 | 3.934E-03 | -26.24% |
| 38 | 0.035 | 0.611 | 5.011E-03 | 3.796E-03 | -24.26% |

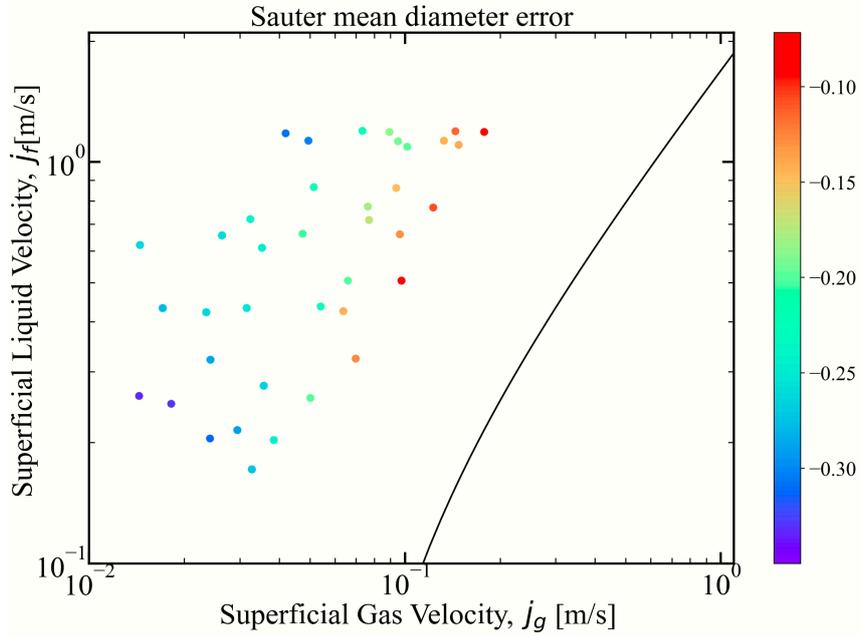

**Fig. 31. MRE map of Sauter mean diameter between BF-GAN and empirical correlation.**

### 4. IAC

In the present study, Zeitoun's empirical correlation [46] was utilized to assess the IAC of images generated by BF-GAN, as outlined in Eq. (24). **Table 17** provides the comparative analysis of the IAC, with a MAMRE of 17.52%. The MRE map for the IAC by BF-GAN is depicted in **Fig. 32**.

$$a_i = 3.24\alpha^{0.757}\left(\frac{g\Delta\rho}{\sigma}\right)^{0.55}\left(\frac{\mu_l}{j_f\rho_l}\right)^{0.1} \qquad (24)$$

**Table 17. Comparative results for IAC [1/m].**

| Num | $j_g$ [m/s] | $j_f$ [m/s] | IAC of empirical correlation | IAC of BF-GAN | MRE |
|---|---|---|---|---|---|
| 1 | 0.029 | 0.215 | 73.705 | 66.302 | -10.04% |
| 2 | 0.024 | 0.322 | 51.264 | 60.246 | 17.52% |
| 3 | 0.070 | 0.324 | 110.146 | 115.977 | 5.29% |
| … | … | … | … | … | … |
| 36 | 0.094 | 0.861 | 72.922 | 84.626 | 16.05% |
| 37 | 0.014 | 0.621 | 23.186 | 14.925 | -35.63% |
| 38 | 0.035 | 0.611 | 45.388 | 43.149 | -4.93% |

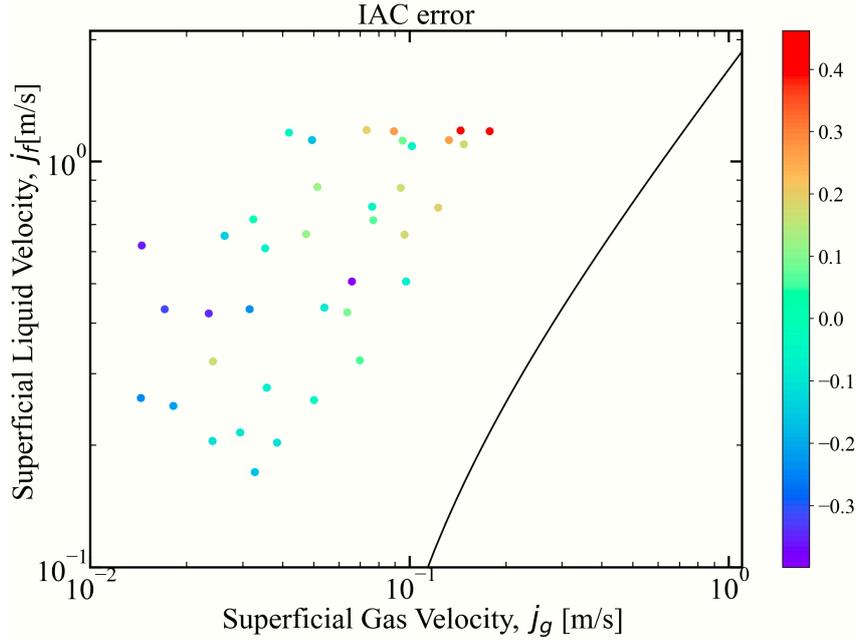

**Fig. 32. MRE map of interfacial area concentration between BF-GAN and empirical correlation.**

**Acknowledgement**

This work was supported by Japan KAKEN 22H02003 and MEXT Innovative Nuclear Research and Development Program Grant Number 21458990.

Wen Zhou appreciates the financial support from the Chinese Scholarship Council (CSC No. 202206340020).

**Declaration of Competing Interest**

The authors declare that they have no known competing financial interests or personal relationships that could have appeared to influence the work reported in this paper.

**Data availability**

The BF-GAN model, bubble detection model, and dataset described in the present study are open-sourced and available in the GitHub repository, accompanied by detailed installation and usage instructions. (https://github.com/zhouzhouwen/BF-GAN)